\documentclass[lettersize,journal]{IEEEtran}
\usepackage{amsmath,amsfonts}
\usepackage{algorithmic}
\usepackage{algorithm}
\usepackage{array}
\usepackage{textcomp}
\usepackage{stfloats}
\usepackage{url}
\usepackage{verbatim}
\usepackage{graphicx}
\usepackage{cite}
\usepackage{subcaption}
\usepackage{caption}
\usepackage{xcolor}
\usepackage{threeparttable}
\usepackage{booktabs}
\usepackage{graphicx}
\usepackage[table]{xcolor}
\usepackage{comment}
\usepackage{multirow} 
\usepackage{makecell}
\usepackage{tablefootnote}

\usepackage{amssymb}

\usepackage[colorlinks=true, citecolor=blue, linkcolor=black]{hyperref}

\begin{document}
	
	\title{DiCoR: Decoupled Referent Disambiguation and Contour Recalibration for Efficient Referring Remote Sensing Image Segmentation}
	
	\author{Ziyang Gao, Zhizhuo Jiang, Jingjing Chang, Yixin Yang, Yuwen Pan, Yong-Qiang Mao, Yu Liu, Hai-Bao Chen
		
		\thanks{This work was supported in part by the National Natural Science Foundation of China under Grant 62401335, Grant 62425117, Grant 62501352; in part by China Postdoctoral Science Foundation under Grant 2025M773484; and in part by the Postdoctoral Innovation Talents Support Program under Grant BX20250413. \textit{(Corresponding authors: Hai-Bao Chen, Yu Liu.)}
			
			Ziyang Gao, Jingjing Chang, Yixin Yang, and Hai-Bao Chen are with the School of Integrated Circuits, School of Information Science and Electronic Engineering, Shanghai Jiao Tong University, Shanghai 200240, China (e-mail: gaoziyang@sjtu.edu.cn; ChangJingjing@sjtu.edu.cn; yangyixin173@sjtu.edu.cn; haibaochen@sjtu.edu.cn).
			
			Zhizhuo Jiang is with the College of Computer Science, Nankai University, Tianjin 300350, China (e-mail: jiangzhizhuo@nankai.edu.cn).
			
			Yuwen Pan is with the Department of Electronic Engineering, Tsinghua Shenzhen International Graduate School, Tsinghua University, Shenzhen 518055, China (e-mail: panyuwen@sz.tsinghua.edu.cn).
			
			Yong-Qiang Mao and Yu Liu are with the Department of Electronic Engineering, Tsinghua University, Beijing 100084, China (e-mail: mao.yongqiang@ieee.org; liuyu\_thu@mail.tsinghua.edu.cn).
			
		}
	}
	
	
	
	\maketitle
	
	\begin{abstract}
		Referring remote sensing image segmentation (RRSIS) aims to delineate targets specified by natural language expressions in remote sensing imagery, providing a flexible paradigm for fine-grained scene interpretation. 
		Existing RRSIS methods generally follow two paradigms: joint fusion segmentation (JFS) and decoupled prompt segmentation (DPS) based on foundation models. 
		JFS supports efficient inference but often yields limited accuracy, since referent localization and mask delineation are optimized under a unified objective. DPS explicitly separates referent localization from mask generation through spatial prompts and powerful segmenters, but usually introduces substantial memory consumption and inference latency.
		To bridge this gap, we propose DiCoR, a decoupled referent disambiguation and contour recalibration framework for RRSIS.
		Built upon the efficient JFS pipeline, DiCoR introduces dedicated optimization mechanisms for two key challenges: identifying the correct referent among ambiguous candidates and recalibrating coarse mask predictions after initial localization.
		Specifically, a disambiguation-aware localization guidance strategy reformulates referent grounding as candidate-level competition by ranking salient candidate regions with adaptive linguistic cues and injecting the resulting localization prior into fused features.
		In addition, a lightweight contour recalibration module predicts residual corrections to coarse logits under localized contour supervision, thereby improving mask delineation with limited computational overhead. Extensive experiments on RefSegRS, RRSIS-D, and RISBench demonstrate the effectiveness of DiCoR.
		On RefSegRS, DiCoR outperforms the competitive JFS method by 4.90\% in mIoU and 2.16\% in gIoU. Efficiency comparisons further show that DiCoR achieves 2.4–6.5× the inference throughput of the evaluated DPS methods, supporting a favorable accuracy-efficiency trade-off.
		The code is available at \url{https://github.com/zyGao1126/DiCoR}.
	\end{abstract}
	
	
	\section{Introduction} \label{sec:introduction}
	Referring Remote Sensing Image Segmentation (RRSIS) has recently emerged as a key task in vision–language understanding for earth observation, enabling precise localization and segmentation of targets described by natural language~\cite{yuan2024rrsis,liu2024rotated,dong2024cross,10816052}. This text-conditioned segmentation paradigm supports high impact applications in complex operational environments, including disaster response, military monitoring, and urban management~\cite{yang2025large,ou2025geopix,lu2025rrsecs,sosa2026enabling}. In practice, RRSIS systems are often required to deliver reliable instance identification and accurate boundary delineation while meeting strict requirements on inference latency and memory footprint to enable real-time assessment~\cite{broni2023real,liu2021light}.
	
	\begin{figure}[!t]
		\captionsetup{belowskip=-10pt}
		\centering
		\includegraphics[width=3.5in]{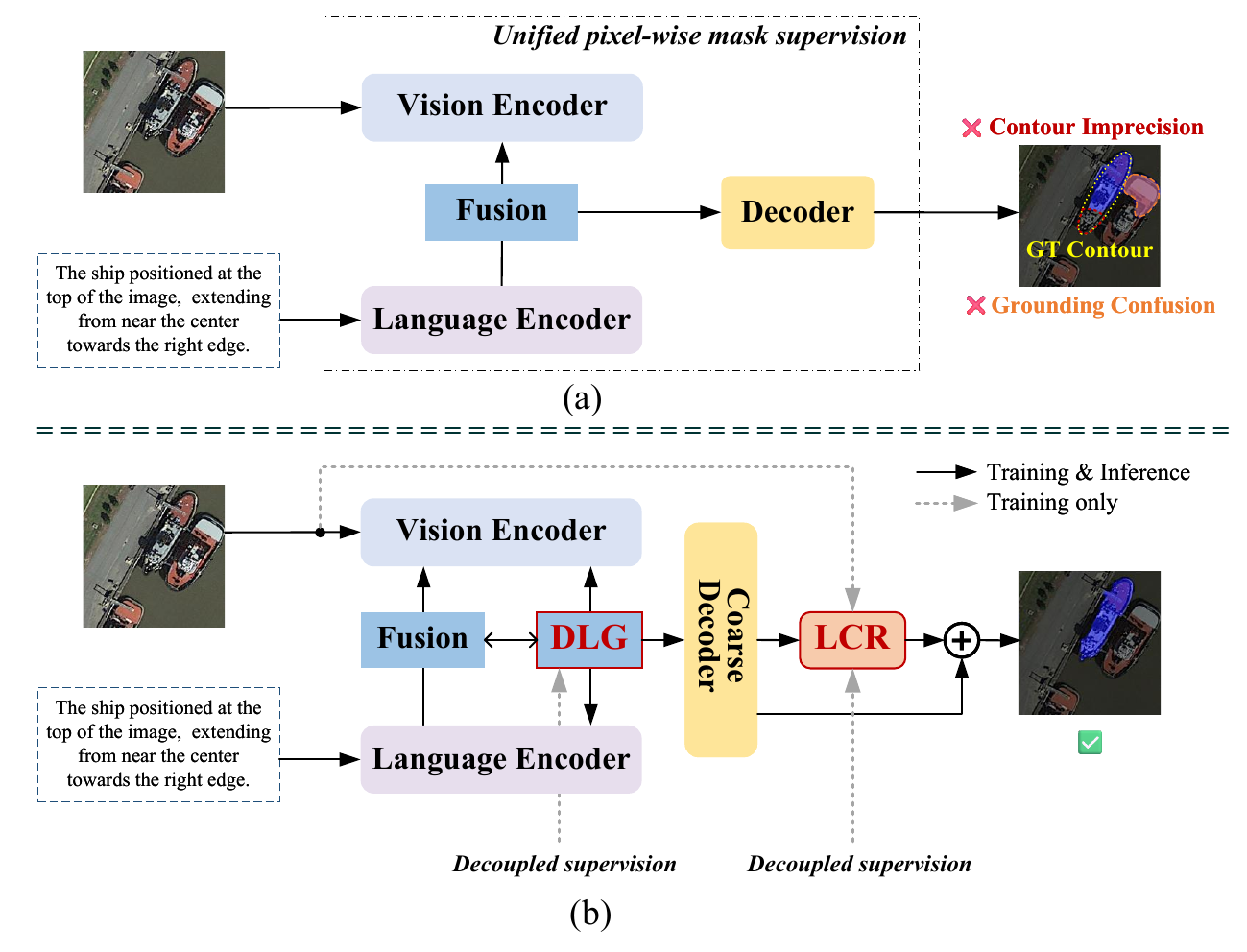}
		\caption{Comparison of conventional JFS pipeline and DiCoR. (a) A conventional JFS pipeline trained under unified pixel-wise mask supervision. (b) DiCoR introduces DLG and LCR, each optimized with dedicated supervision for referent disambiguation and contour recalibration. }
		\label{fig:introduction_framework}
	\end{figure}

	
	Existing RRSIS methods can be broadly grouped into two paradigms: Joint Fusion Segmentation (JFS)~\cite{yuan2024rrsis,liu2024rotated,dong2024cross,10816052,pan2024rethinking,liu2025cadformer,ma2025lscf,li2025scale,zhang2026stdnet,yang2025referring,sun2026crobim,zhang2026multiscale} and Decoupled Prompt Segmentation (DPS)~\cite{chen2025rsrefseg,chen2025rsrefseg2,li2025segearth,xin2025segearth,rong2025customized}. 
	JFS performs referring segmentation in an end-to-end manner by injecting linguistic features into the visual backbone via cross-modal fusion, thereby jointly learning target grounding and mask delineation from fused representations. This unified design is generally inference-efficient, as the target mask can be generated in a single forward pass without invoking an external segmenter. However, JFS is commonly optimized with pixel-wise losses that jointly supervise localization and segmentation accuracy, which may bias training toward coarse localization errors that incur larger penalties while providing weaker correction for fine boundary discrepancies after the target has been approximately localized. 
	This limitation is particularly pronounced in remote sensing imagery, where small objects are prevalent and minor contour deviations can substantially affect segmentation quality.
	
	In contrast, the DPS paradigm decouples target grounding from mask generation by first deriving intermediate spatial prompts and then feeding these prompts into strong segmentation foundation models to produce the final mask.
	By externalizing “where to segment” as a prompt, DPS provides a clearer interface between localization and mask synthesis and often achieves accurate mask delineation. 
	However, foundation-model-based pipelines typically incur substantial inference time overhead, resulting in increased memory footprint and latency that can hinder practical deployment. 
	Moreover, such pipelines can be sensitive to remote sensing domain characteristics, such as sensor modality variations and complex background textures, making domain shift a persistent challenge.
	
	These observations raise a central question: \textit{can we preserve the efficiency of JFS while incorporating the grounding reliability and contour fidelity offered by DPS?} 
	To this end, we propose DiCoR, a decoupled referent disambiguation and contour recalibration framework for RRSIS. DiCoR is motivated by the decoupling philosophy behind DPS, yet implements it with lightweight modules and task-specific supervision instead of relying on a heavy foundation model during inference.
	 We next explain how this objective is achieved through referent disambiguation and contour recalibration.
	
	
	\textbf{(1) Referent ambiguity under distractors and competing cues.} A central challenge in RRSIS is to identify the target referent among distractor candidates that share similar appearance, geometry, or spatial context. This challenge is further amplified when an expression contains multiple cues, such as location, size, and attributes, whose reliability varies across samples. Conventional JFS methods couple referent grounding with mask prediction under a unified objective, which does not explicitly resolve the competition among plausible same-image candidates.
	DiCoR addresses this issue by reformulating referent grounding as a candidate competition problem, where the model is encouraged to distinguish competing high-response regions and concentrate evidence on the true referent. From the representation perspective, we introduce a Disambiguation-aware Localization Guidance (DLG) strategy that constructs a compact set of candidate regions from the intermediate response map and ranks them with candidate-conditioned linguistic and geometric evidence. 
	From the supervision perspective, DLG is trained with a complementary localization-and-ranking objective, where distractors extracted by offline SAM3~\cite{carion2025sam} are used to guide the intermediate response map, and a ranking loss encourages the model to select the ground-truth referent over competing candidates.
	
	\textbf{(2) Contour imprecision under coarse localization.} Even when the referent has been approximately localized, accurate mask delineation remains difficult. This is because standard pixel-wise supervision is dominated by localization errors, leaving much weaker signals for correcting the remaining contour deviations. As a result, boundary errors are often insufficiently corrected in JFS models. 
	To address this issue, DiCoR introduces a Lightweight Contour Recalibration (LCR) module after the coarse decoder and trains it in a decoupled manner. Taking the input image and the coarse prediction as input, LCR learns residual corrections to the coarse logits under localized contour supervision that emphasizes uncertain boundary regions, rather than re-segmenting the target from scratch. To improve robustness under realistic coarse priors, coarse predictions from multiple checkpoints are collected and filtered by localization quality, then further diversified with morphological perturbations.

	Fig.~\ref{fig:introduction_framework} provides a conceptual summary of DiCoR. By moving beyond the unified pixel-wise supervision used in conventional JFS, DiCoR adopts a lightweight decoupled design centered on the two key challenges of referent disambiguation and contour recalibration, while maintaining efficient inference. To summarize, the main contributions of this paper are as follows:
	
	\begin{itemize}
		\item We propose DiCoR, a decoupled referent disambiguation and contour recalibration framework for efficient RRSIS, which improves both grounding reliability and mask delineation quality within a lightweight and efficient inference pipeline.
		\item We introduce a disambiguation-aware localization guidance strategy that reformulates referent grounding in RRSIS as a candidate competition problem, enabling the model to identify the correct referent through candidate-level discrimination and explicit ranking supervision.
		\item We introduce a lightweight contour recalibration module, inspired by the decoupling philosophy of prompt-based methods, which improves boundary delineation through residual correction with limited inference overhead.
		\item Extensive experiments on public RRSIS benchmarks demonstrate that DiCoR achieves a favorable trade-off between segmentation accuracy and deployment efficiency.
	\end{itemize}	

	The remainder of this paper is organized as follows. Section~\ref{sec:related_work} reviews related studies on referring image segmentation and referring remote sensing image segmentation. Section~\ref{sec:method} presents the proposed DiCoR framework and details its referent disambiguation and contour recalibration components. Section~\ref{sec:experiments} reports comprehensive experiments on public RRSIS benchmarks and analyzes the effectiveness of the proposed modules. Finally, Section~\ref{sec:conclusion} concludes the paper.
	
	\begin{figure*}[!ht]
		\centering
		\includegraphics[width=\textwidth]{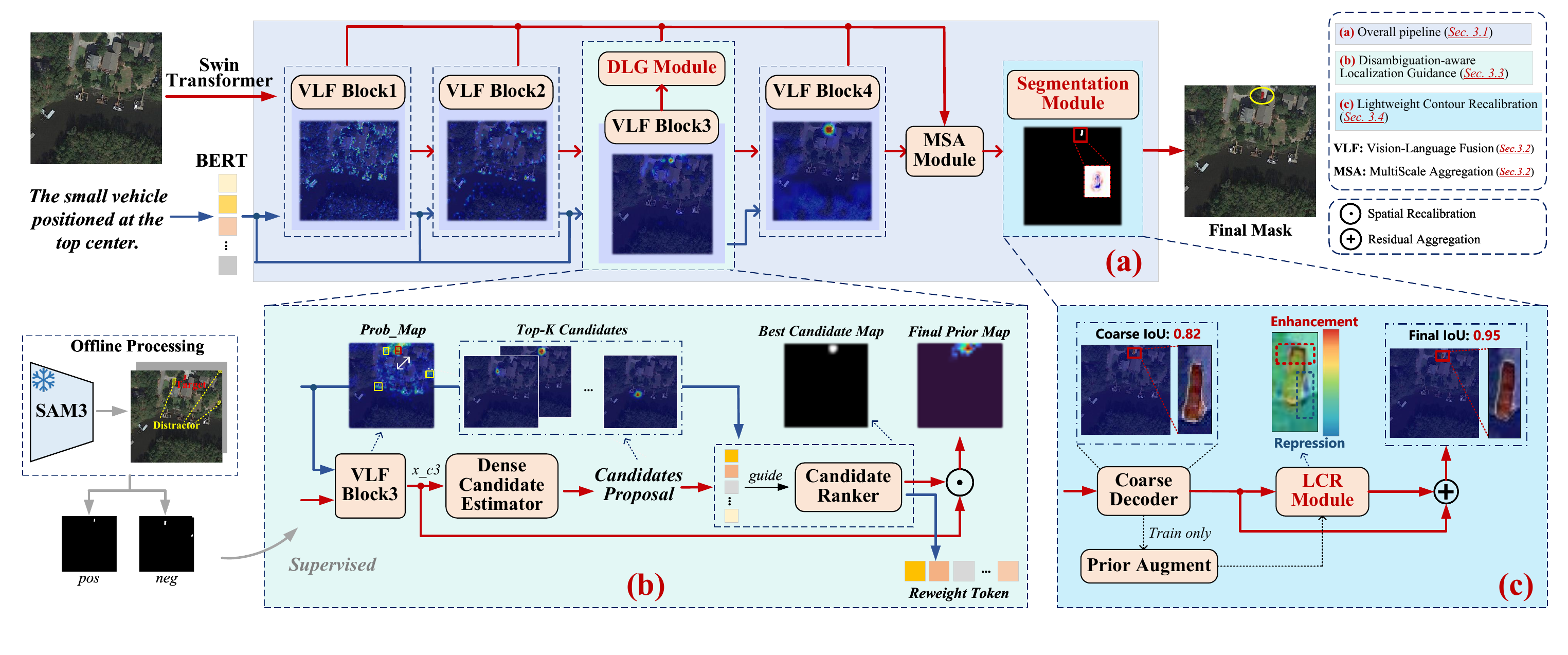}
		\caption{Overview of the proposed DiCoR framework, including (a) the overall pipeline built upon a JFS backbone, (b) disambiguation-aware localization guidance for referent disambiguation, and (c) lightweight contour recalibration for boundary correction.}
		\label{fig:fig_overview}
	\end{figure*}
	
	\section{Related Work} \label{sec:related_work}
	\subsection{Referring Image Segmentation} \label{sec:ris}
	Referring image segmentation (RIS) aims to segment image regions specified by natural-language expressions. Early RIS methods~\cite{hu2016segmentation,li2018referring,margffoy2018dynamic,ye2019cross, hu2020bi} typically combined CNN-based visual encoders with RNN/LSTM language encoders, followed by late multimodal fusion for pixel-wise prediction. Subsequent transformer-based approaches introduced stronger global context modeling and more flexible vision-language interaction~\cite{luo2020multi,feng2021encoder,ding2021vision, kim2022restr}, thereby mitigating the locality limitations of earlier CNN–RNN pipelines. Building on this trend, LAVT\cite{yang2022lavt} advanced RIS by injecting linguistic information into the intermediate stages of the visual encoder, enabling language-aware visual representation learning throughout the encoding process rather than only after unimodal feature extraction. This design substantially influenced later RIS research, with methods such as SLViT\cite{ouyang2023slvit} and CARIS\cite{liu2023caris} further improving scale-aware interaction and context-aware alignment. 
	More recently, RIS has increasingly benefited from large pretrained models\cite{wang2022cris,kim2024extending,shang2024prompt,lai2024lisa}, particularly through CLIP-based alignment transfer~\cite{xu2023bridging,yu2023zeroshot} and prompt-driven SAM frameworks~\cite{zhang2024evf,ito2025feature}, which further improve semantic understanding and mask generation quality. 
	Overall, RIS in natural images has evolved toward stronger cross-modal alignment, finer interaction, and more effective exploitation of pre-trained knowledge. However, these methods are largely developed for natural-image scenarios and therefore do not directly address the small-object density, arbitrary orientations, long-range spatial relations, and referential ambiguity that characterize remote sensing imagery. Related boundary-aware segmentation studies have further shown that contour-sensitive modeling and boundary displacement correction can improve mask delineation near object edges~\cite{takikawa2019gated,yuan2020segfix,cheng2021boundary}, which also provides useful motivation for contour recalibration in RRSIS.
	
	\subsection{Referring Remote Sensing Image Segmentation} \label{sec:rrsis}
	Referring image segmentation was first introduced into the remote sensing domain by \cite{yuan2024rrsis}, which established a dedicated benchmark and stimulated subsequent research in this area. Most existing studies follow the joint fusion segmentation (JFS) paradigm introduced above, where multimodal encoding, cross-modal interaction, and pixel-level decoding are optimized within a unified end-to-end network. Early JFS methods mainly focused on modeling characteristics specific to remote sensing imagery, such as large-scale variation and orientation diversity \cite{liu2024rotated}, while later studies progressively improve multimodal representation through finer image–text alignment, stronger bidirectional interaction, and more effective multi-level feature aggregation \cite{10816052, dong2024cross, li2025scale, zhang2026stdnet}. More recent work has further explored longer-range semantic guidance, dual alignment, uncertainty-aware modeling, fine-grained decoding, and stronger collaborative encoding strategies \cite{pan2024rethinking,ma2025lscf,liu2025cadformer,sun2026crobim,zhang2026multiscale,yang2025referring}. In particular, CroBIM-U~\cite{sun2026crobim} uses a shared pixel-wise uncertainty map to guide cross-modal language modulation and local refinement within a unified training framework. However, this shared uncertainty formulation does not explicitly distinguish referent ambiguity from residual contour errors as separate learning problems with dedicated supervision.
	
	Meanwhile, recent advances in large pretrained vision–language and promptable segmentation models have opened another important direction for RRSIS. In these methods, which we term decoupled prompt segmentation (DPS), multimodal cues are converted into explicit spatial prompts by multimodal representation or grounding models~\cite{radford2021learning,li2022glip,liu2024grounding,liu2024remoteclip}. Strong pretrained segmenters~\cite{kirillov2023segment,zou2023seem,ke2023hqsam,ravi2025sam2} are then adopted to generate the final masks. Representative studies have developed prompting pipelines built on foundation models, customized promptable segmentation frameworks, and segmentation schemes that combine coarse localization with subsequent mask refinement~\cite{chen2025rsrefseg,rong2025customized,li2025semantic,chen2025rsrefseg2}. Recent studies on large models further broaden this direction to more diverse query forms and more complex reasoning processes~\cite{li2025segearth,xin2025segearth}. Although these approaches benefit from the rich prior knowledge and strong mask generation capability of foundation models, their reliance on external heavy segmenters and staged inference pipelines typically leads to higher computational and memory overhead, while also making overall performance more sensitive to prompt quality and cross-domain adaptation. Our method is conceptually aligned with the decoupling motivation of this line of research, but realizes it within a lightweight implicit fusion pipeline rather than through a heavy external segmenter.

	\section{Methodology} \label{sec:method}
	\subsection{Overview} \label{sec:overview}
	We first present an overview of DiCoR in Fig.~\ref{fig:fig_overview}. Built on an efficient end-to-end fusion pipeline, DiCoR is designed to improve referent disambiguation and contour recalibration in response to two persistent challenges in RRSIS. Specifically, the input image is first encoded by Swin Transformer backbone~\cite{liu2021swin}, and the referring expression is encoded by BERT~\cite{devlin2019bert}. The two streams are progressively integrated through four vision-language fusion (VLF) blocks to produce multi-level multimodal representations. These features are then aggregated by a multi-scale aggregation (MSA) module and decoded to generate a coarse segmentation prediction, which is further refined to obtain the final mask.
	
	To improve referent localization, DiCoR introduces a disambiguation-aware localization guidance (DLG) module, which formulates referent grounding as a candidate competition process. This design allows the model to resolve ambiguity among same-image candidates and focus on the top-ranked referent. Starting from the fused representation after the third VLF block, the DLG module first predicts an intermediate response map and organizes salient responses into a compact set of candidate regions. These candidates are then assessed using candidate-aware language cues, where token responses are adaptively reweighted according to candidate-level visual evidence. The selected candidate is subsequently transformed into a localization prior and injected back into the same fused representation through residual spatial recalibration, thereby guiding subsequent fusion and decoding toward the true referent. During training, same-image distractors mined offline by SAM3 model provide explicit competitive supervision, encouraging the model to preserve a discriminative candidate structure and distinguish the ground-truth referent from visually confusing competitors.
	
	For contour recalibration, DiCoR further employs a lightweight contour recalibration (LCR) module after the coarse decoder. Taking the input image and the coarse prediction as input, LCR predicts a residual correction to the coarse logits, so that the final mask is improved through localized boundary adjustment rather than full re-segmentation. Accordingly, LCR adopts a lightweight encoder-decoder architecture with mirrored skip connections, enabling local contour correction while preserving fine spatial details. For training data construction, we collect coarse predictions from multiple intermediate checkpoints, retain samples with sufficiently reliable localization, and further diversify them with morphological perturbations to cover richer boundary deviations. For training supervision, LCR is optimized with a region-aware objective that emphasizes uncertain boundary regions while maintaining supervision over the predicted foreground and its neighborhood.

	\subsection{Multi-Scale Joint Fusion Backbone} \label{sec:backbone}
	Given a remote sensing image $I \in \mathbb{R}^{3 \times H \times W}$ and a referring expression $T=(w_1,\ldots,w_n)$, the visual encoder produces a hierarchy of visual features at four scales, denoted by $\{V_l\}_{l=1}^{4}$, while the language encoder converts the expression into token embeddings $L=[l_1,\ldots,l_n]^\top$. Here, $l_n$ denotes the embedding of the $n$-th token. At each fusion stage, a language-guided attention module incorporates linguistic information from $L$ into the corresponding visual feature $V_l$, yielding multi-level vision-language features $\{X_l\}_{l=1}^{4}$. These representations preserve complementary cues across resolutions, with shallow features emphasizing local spatial detail and deeper features capturing richer semantic context. The multi-scale features are subsequently aggregated through MSA module and decoded to produce the prediction mask $P$.
	
\begin{figure}[htb]
		\captionsetup{belowskip=-10pt}
		\centering
		\includegraphics[width=3.5in]{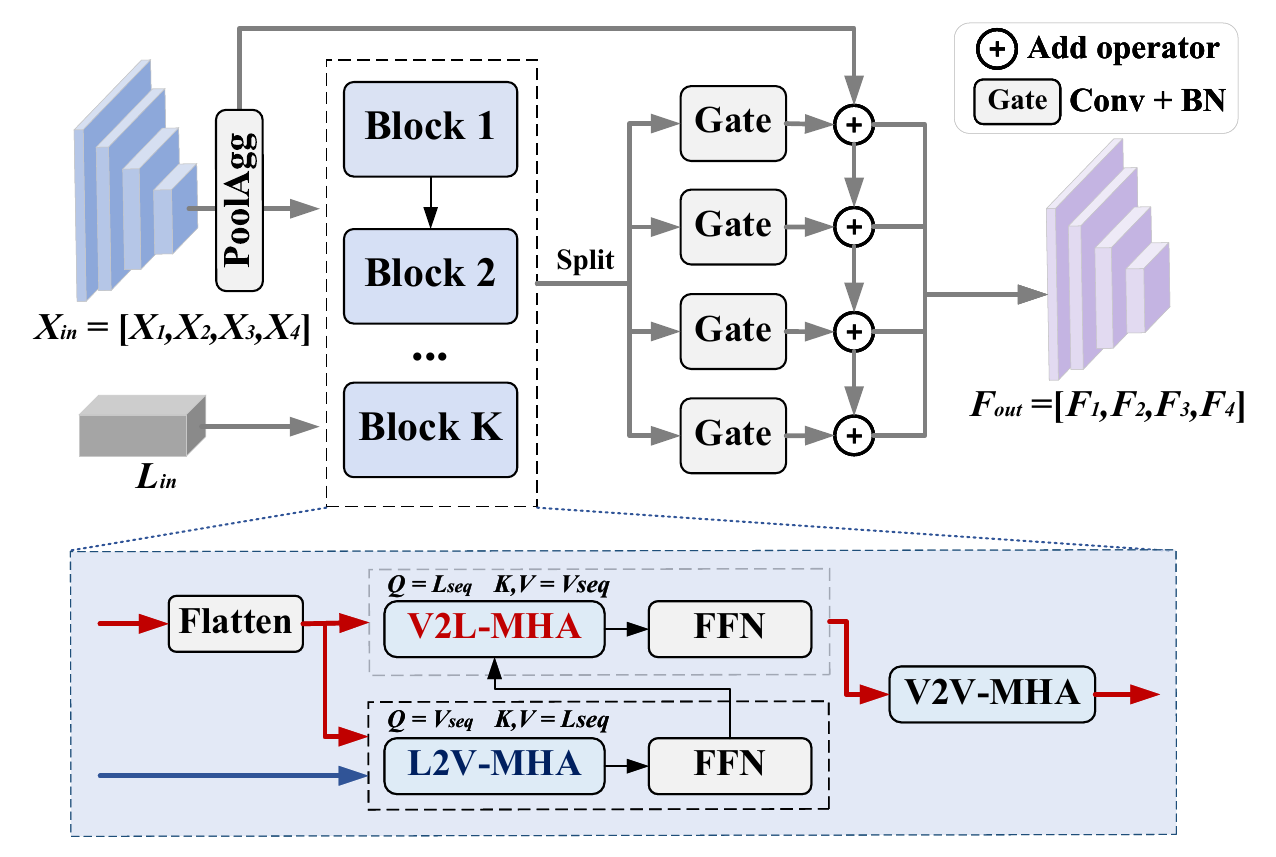}
		\caption{ Structure of the MSA module. }
		\label{fig:lvmsf}
	\end{figure}

	To strengthen cross-scale interaction, bidirectional vision-language attention is introduced into the MSA module. As illustrated in Fig.~\ref{fig:lvmsf}, multimodal features from four encoder stages are first aligned by pyramid pooling and flattened into a unified visual sequence. Different from conventional aggregation methods that mainly inject textual semantics into visual features, the proposed interaction scheme further uses visual evidence to update linguistic representations. Concretely, L2V multi-head attention refines the visual sequence under language guidance, whereas V2L multi-head attention recalibrates the language sequence according to the visual content. The updated visual representation is subsequently processed by visual self-attention to model long-range spatial dependencies and global contextual relations. Finally, the aggregated representation is decomposed back into four scales and fused into the original feature hierarchy through scale-aware gating, allowing information to propagate adaptively across resolutions.
	
	\subsection{Disambiguation-aware Localization Guidance} \label{sec:dlg}
\begin{figure}[htb]
		\centering
		\includegraphics[width=3.5in]{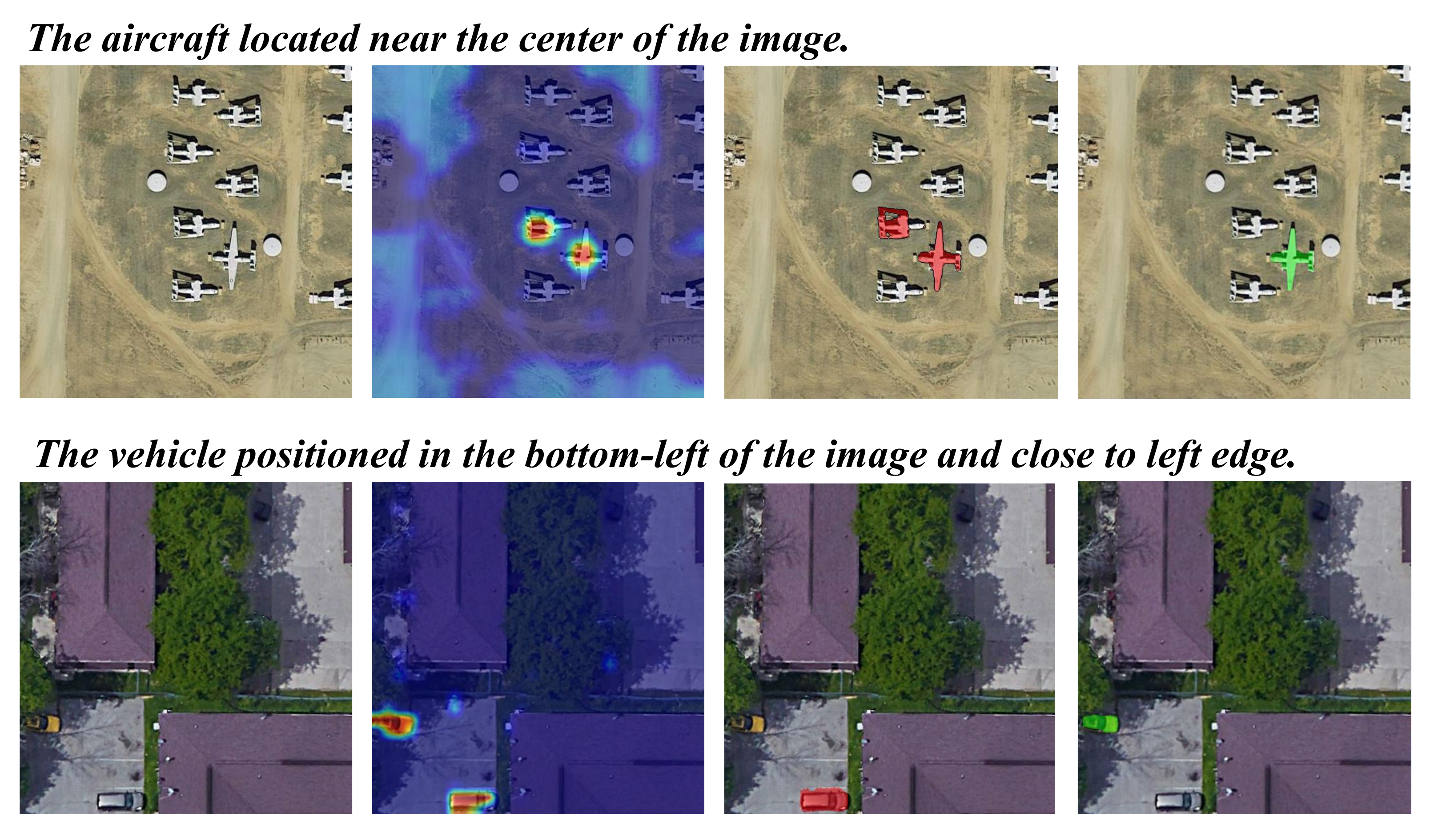}
		\caption{Failure cases of the JFS backbone. From left to right are the input image, the activation visualization of $X_3$, the predicted mask, and the ground-truth mask. }
		\label{fig:ambiguityCase}
	\end{figure}

	Although the joint fusion backbone provides an efficient basis for RRSIS, it remains vulnerable to ambiguous grounding when multiple same-image instances share similar appearance or spatial layout. As shown in Fig.~\ref{fig:ambiguityCase}, the fused feature $X_3$ already activates the ground-truth referent, but comparable responses also appear on confusing distractors. This suggests that the failure is not simply caused by missing referent evidence. Instead, multiple plausible regions are often activated simultaneously, while the model lacks an explicit mechanism to resolve their competition. Since standard pixel-wise supervision mainly enforces foreground-background separation, it provides limited guidance for distinguishing among high response candidate instances. The resulting ambiguity can therefore propagate to the decoder and lead to incorrect masks.
	
\begin{figure}[htb]
		\centering
		\includegraphics[width=3.5in]{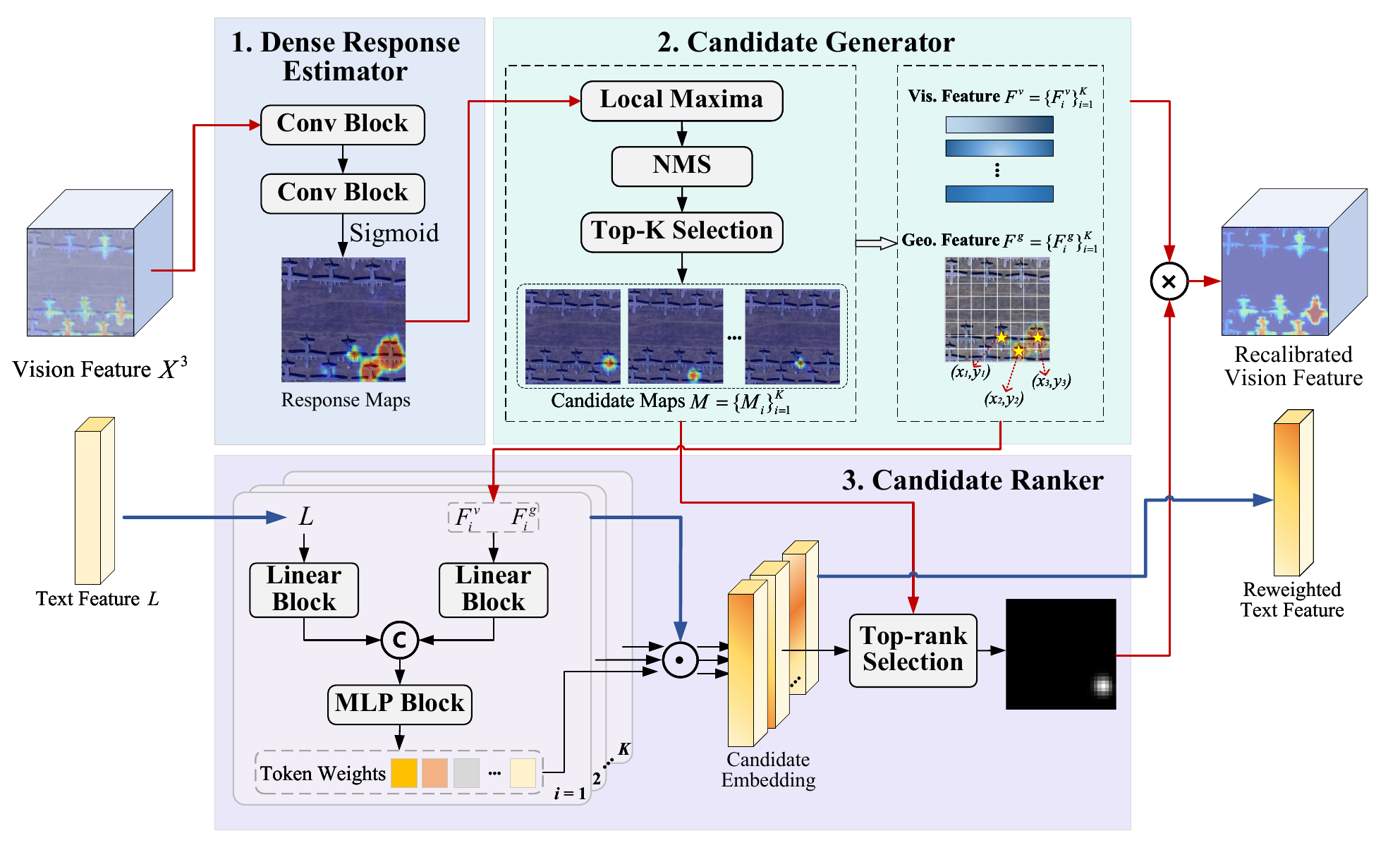}
		\caption{ Architecture of the proposed DLG module. DLG produces a candidate-specific localization prior and a reweighted text feature, and then feeds the selected prior back into $X_3$ for spatial recalibration.}
		\label{fig:locate}
	\end{figure}

	To make this competition explicit, DLG module is introduced to reformulate referent grounding as candidate-level competition. As illustrated in Fig.~\ref{fig:locate}, DLG contains three stages: a dense response estimator, a candidate generator, and a candidate ranker. Given the third-stage fused feature $X_3$ and the language feature $L$, the dense response estimator first predicts a response map
	\begin{equation}
		R = \Phi_{\mathrm{resp}}(X_3), \quad
		R \in [0,1]^{H_3 \times W_3}
	\end{equation}
	where $\Phi_{\mathrm{resp}}$ is implemented by two lightweight Conv-Norm-GELU blocks followed by a sigmoid activation. The response map $R$ captures the spatial distribution of potential referent cues in the deep multimodal representation.
	
	The candidate generator then extracts a compact set of candidate regions from $R$. To this end, local response peaks are filtered through non-maximum suppression, and the top-$K$ remaining locations ${p_i}_{i=1}^{K}$ are retained. These selected peaks are subsequently converted into localized candidate supports through Gaussian-gated response aggregation
	\begin{equation}
		C_i(u) = R(u) \cdot \exp\left(-\frac{\|u - p_i\|_2^2}{2\sigma_c^2}\right), \quad i = 1, \ldots, K,
	\end{equation}
	where $u$ denotes the spatial coordinate and $\sigma_c$ controls the support range. After normalization, each candidate support is used to pool a visual feature $v_i$ from $X_3$. In parallel, a geometric feature $g_i$ is constructed from the spatial distribution, including the normalized center, effective area, and spatial dispersion. Together, $v_i$ and $g_i$ define the candidate representation for subsequent ranking.

	Next, the candidate ranker estimates referent confidence for each candidate. Since different regions may rely on different linguistic cues for disambiguation, token importance should be estimated in a candidate-adaptive manner. For candidate $i$, the concatenated visual and geometric features interact with the language features through lightweight MLP projections followed by a routing network. We denote this mapping by $\Phi_{\mathrm{rt}}$, which produces normalized token weights:
	\begin{equation}
		\omega_i = \Phi_{\mathrm{rt}}(v_i,g_i,L), \quad
		\omega_i \in \mathbb{R}^{N},
	\end{equation}
	which are then used to reweight the token-level language features to produce a candidate-specific text representation:
	\begin{equation}
		\tilde{l}_i = \sum_{n=1}^N \omega_{i,n} l_n.
	\end{equation}
	
	The referent confidence score $s_i$ of each candidate is then computed by jointly considering semantic similarity and geometric consistency:
	\begin{equation}
		s_i = \langle v_i, \tilde{l}_i \rangle + \lambda_g \Phi_g\left([g_i; \tilde{l}_i]\right)
	\end{equation}
	where $\langle \cdot,\cdot \rangle$ denotes cosine similarity computation, $\Phi_g(\cdot)$ is implemented as a lightweight MLP, and $\lambda_g$ balances the contribution of geometric evidence. These scores establish an explicit ranking among the candidates, with the highest-confidence candidate $i^*$ selected as the predicted referent. The corresponding support is normalized by its spatial maximum and then sharpened through an element-wise power transformation to emphasize high-response regions. The resulting support $\bar{C}_{i^*}$ is injected back into $X_3$ through residual spatial recalibration:
	\begin{equation}
		\tilde{X}_3 = X_3 \odot \left(1 + \gamma_{\mathrm{dlg}} \bar{C}_{i^*}\right),
	\end{equation}
	where $\gamma_{\mathrm{dlg}}$ controls the guidance strength and $\odot$ denotes broadcasting along the channel dimension. To focus DLG on resolving ambiguity among small and competing instances in remote sensing scenes, we train it on samples with ground-truth foreground area ratios in $(0,\tau)$ and apply it at inference only if the coarse-mask foreground ratio is below the same threshold. In this way, the recalibrated feature $\tilde{X}_3$ guides the decoder toward the resolved referred instance.
	
	To supervise the DLG module effectively, it is necessary to jointly ensure accurate candidate ranking and reliable candidate generation. As the candidate generation process relies on non-differentiable discrete operations, directly optimizing candidate quality through gradient-based learning is non-trivial. Consequently, the quality of the response map $R$ becomes the primary factor governing candidate reliability.
	
	We therefore introduce two complementary objectives: response supervision for stable candidate generation and ranking supervision for explicit referent selection. Specifically, response supervision encourages $R$ to produce strong activations over potential referent regions. Directly applying ground-truth masks as positive supervision is inconsistent with the role of deep multimodal features, which are expected to highlight candidate evidence regions rather than reconstruct rigid target extents. To address this, we partition the response space into three regions, including the target positive foreground $\mathcal{P}$, the clean background $\mathcal{B}$, and the hard instance background $\mathcal{H}$, where $\mathcal{P}$ is defined as a compact region centered at the ground-truth target, $\mathcal{B}$ consists of regions sufficiently distant from the target, and $\mathcal{H}$ captures hard distractor instances. The response supervision loss is defined as
	\begin{equation}
		\mathcal{L}_{\mathrm{resp}} 
		=
		\mathcal{L}_{\mathrm{bce}}(R_{\mathcal{P}}, \mathbf{1}) + \mathcal{L}_{\mathrm{bce}}(R_{\mathcal{B}}, \mathbf{0}) +
		\lambda_h \mathcal{L}_{\mathrm{bce}}(R_{\mathcal{H}}, \mathbf{0})
	\end{equation}
	here, $\mathcal{L}_{\mathrm{bce}}(\cdot,\cdot)$ denotes the standard binary cross-entropy loss, and each term is averaged over the corresponding region. $R_{\mathcal{P}}$, $R_{\mathcal{B}}$, $R_{\mathcal{H}}$ denote the response values sampled from region $\mathcal{P}$, $\mathcal{B}$ and $\mathcal{H}$ respectively. 
	It should be noted that $\mathcal{H}$ is constructed via offline mining using a frozen SAM3 model. Since SAM3 is employed exclusively for offline distractor mining, its masks provide auxiliary hard-negative supervision without serving as model inputs or final segmentation targets.
	Candidate object proposals with high confidence but low overlap with the ground-truth mask are retained as distractors and aggregated into a proposal set to form $\mathcal{H}$, thereby introducing structured hard negatives that explicitly model intra-image ambiguity.
	
	To resolve ambiguity among competing candidates, we further impose ranking supervision via a cross-entropy loss. Let $y$ denote the ground-truth candidate index, the ranking loss is defined as
	\begin{equation}
		\mathcal{L}_{\mathrm{rank}}
		= -\log \frac{\exp(s_y)}{\sum_{i=1}^{K} \exp(s_i)}.
	\end{equation}
	This objective enforces explicit competition among candidates and promotes correct referent selection. The overall DLG objective is formulated as
	\begin{equation}
		\mathcal{L}_{\mathrm{DLG}}
		= \lambda_{\mathrm{resp}}\mathcal{L}_{\mathrm{resp}}
		+ \lambda_{\mathrm{rank}}\mathcal{L}_{\mathrm{rank}},
	\end{equation}
	where $\lambda_{\mathrm{resp}}$ and $\lambda_{\mathrm{rank}}$ control the relative contributions of response estimation and candidate ranking, yielding complementary supervision during training.

	\subsection{Lightweight Contour Recalibration} \label{sec:lcr}
	Although DLG improves referent localization by resolving ambiguity through candidate-level disambiguation, accurate mask prediction still depends on the decoder's ability to delineate object contours. In conventional JFS frameworks, localization and contour delineation are optimized jointly under a unified mask objective. Once the referent has been approximately grounded, boundary deviations tend to induce weaker gradients than residual localization errors under the same objective, leading to insufficient optimization of fine-grained contour delineation. To address this optimization imbalance, we introduce a decoupled lightweight contour recalibration module, termed LCR, which explicitly decouples contour correction from coarse localization via residual recalibration.
	
	\begin{figure}[htb]
		\captionsetup{belowskip=-10pt}
		\centering
		\includegraphics[width=3.5in]{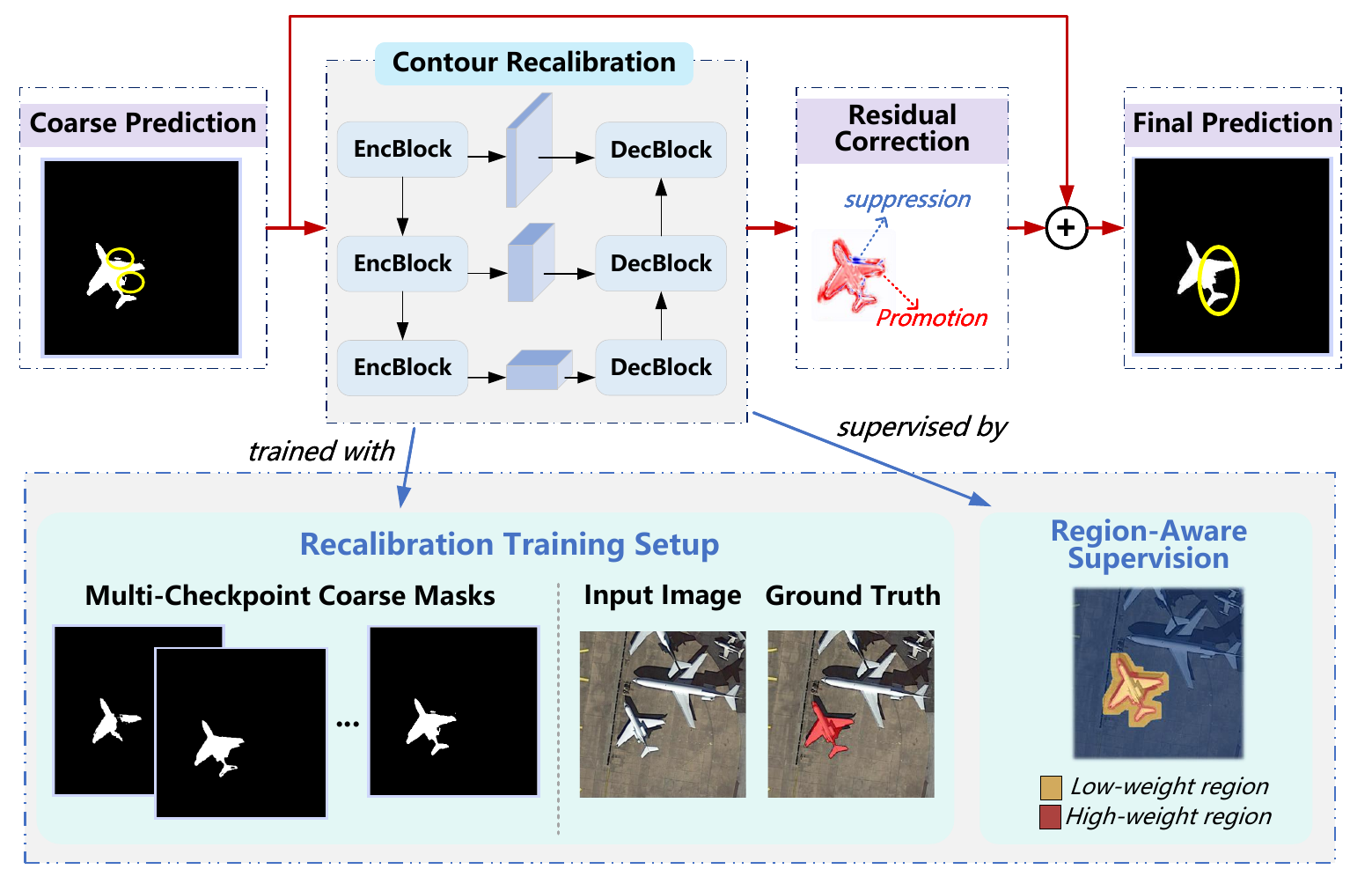}
		\caption{ Structure of the LCR module. }
		\label{fig:refiner}
	\end{figure}
	
	Let $Z$ denote the coarse segmentation logits produced by the decoder and $P = \sigma(Z)$ denote the corresponding mask prediction through sigmoid activation, the LCR module takes the input image $I$ and $P$ as guidance to estimate a residual correction $\Delta Z$ in the logit domain:
	\begin{equation}
		\Delta Z = \Phi_{\mathrm{lcr}}\left(I \oplus P\right),
	\end{equation}
	where $\oplus$ denotes channel-wise concatenation and $\Phi_{\mathrm{lcr}}$ denotes the lightweight recalibration network. As illustrated in Fig.~\ref{fig:refiner}, $\Phi_{\mathrm{lcr}}$ adopts a compact encoder–decoder architecture composed of convolutional layers, normalization and ReLU activation. Skip connections are introduced between mirrored stages to preserve fine spatial structures during recalibration. The recalibrated logits $\tilde{Z}$ and the final prediction $\tilde{P}$ are obtained as
	\begin{equation}
		\tilde{Z} = Z + \gamma_{\mathrm{lcr}}\Delta Z, \quad
		\tilde{P} = \sigma(\tilde{Z})
	\end{equation}
	where $\gamma_{\mathrm{lcr}}$ denotes the residual combination coefficient. This formulation encourages LCR to perform localized contour correction, while building on the coarse localization provided by $Z$.
	
	To focus optimization on residual contour errors, LCR adopts a region-aware loss function instead of uniform pixel-wise supervision. As illustrated in Fig.~\ref{fig:refiner}, the weight map $W(u)$ is constructed from the coarse prediction to emphasize pixels near predicted contours while downweighting distant background regions with confident predictions. Let $G$ denote the ground-truth binary mask, the training objective is defined as
	\begin{equation}
		\mathcal{L}_{\mathrm{LCR}}
		=
		\lambda_{\mathrm{ce}}\mathcal{L}_{\mathrm{rce}}
		+
		\lambda_{\mathrm{dice}}\mathcal{L}_{\mathrm{rdice}},
	\end{equation}
	where $\mathcal{L}_{\mathrm{rce}}$ and $\mathcal{L}_{\mathrm{rdice}}$ denote the region-aware cross-entropy and Dice losses, which are commonly used in RRSIS tasks to balance pixel accuracy and region overlap~\cite{chen2025rsrefseg2,ma2025lscf,10816052}. The coefficients $\lambda_{\mathrm{ce}}$ and $\lambda_{\mathrm{dice}}$ balance the two loss terms. Specifically, $\mathcal{L}_{\mathrm{rce}}$ is defined as:
	\begin{equation}
		\mathcal{L}_{\mathrm{rce}}
		=
		\frac{1}{\sum_{u \in \Omega} W(u)}
		\sum_{u \in \Omega}
		W(u)\operatorname{CE}\left(\tilde{P}(u),G(u)\right),
	\end{equation}
	where $\Omega$ denotes the spatial domain and $u$ indexes pixel locations. The weight map $W(u)$ modulates the contribution of each pixel by assigning larger values to contour-adjacent regions and smaller values to regions with confident predictions. Similarly, $\mathcal{L}_{\mathrm{rdice}}$ is computed as:
	\begin{equation}
		\mathcal{L}_{\mathrm{rdice}}
		=
		\operatorname{Dice}
		\left(
		\{W(u)\tilde{P}(u)\}_{u \in \Omega},
		\{W(u)G(u)\}_{u \in \Omega}
		\right).
	\end{equation}
	Together, these losses define a spatially selective supervision that enables effective contour recalibration through residual correction based on coarse prior, providing an efficient alternative to foundation-model-based refinement pipelines.
	
	\subsection{Decoupled Training Strategy} \label{sec:training_strategy}
	This section presents the decoupled training strategy for optimizing DiCoR. Following existing JFS methods~\cite{10816052,ma2025lscf}, the vision-language fusion backbone and the coarse decoder are first trained with the standard segmentation loss. To provide the auxiliary modules with a broader and more realistic training distribution, intermediate checkpoints from different training stages are subsequently sampled to construct diverse pretraining data. Specifically, third-level fused features are extracted for DLG pretraining, while coarse decoder logits are collected as foreground priors for LCR. For LCR pretraining, only samples with sufficiently reliable coarse localization are retained. This filtering removes severely mislocalized predictions that would otherwise introduce inconsistent correction targets and interfere with LCR optimization. To further enrich boundary variations and improve the generalization ability of contour recalibration, the retained coarse priors are augmented with morphological perturbations, including dilation and erosion.
	DLG and LCR are trained independently while the coarse segmentation model remains frozen. At inference time, DiCoR adds both the lightweight modules to a standard JFS pipeline, thus avoiding the heavy computational cost of foundation-model-based segmentation.
	
	\section{Experiments} \label{sec:experiments}
	\subsection{Datasets and Metrics} \label{sec:datasets_metrics}
	\noindent\textbf{Datasets.}
	We evaluate DiCoR on three public referring remote sensing image segmentation benchmarks, including RefSegRS~\cite{yuan2024rrsis}, RRSIS-D~\cite{liu2024rotated}, and RISBench~\cite{dong2024cross}. RefSegRS contains 4,420 image-expression-mask triplets collected from 285 remote sensing scenes, covering 14 object categories. The dataset is split into 2,172 training samples, 431 validation samples, and 1,817 test samples. RRSIS-D provides a larger benchmark with 17,402 annotated samples from 20 categories, including 12,181 training samples, 1,740 validation samples, and 3,481 test samples. RISBench further increases the scale, scene diversity, and referring complexity of the task, providing a challenging testbed for evaluating model robustness. These datasets exhibit substantial differences in object scale, scene composition, and semantic coverage, and therefore provide a comprehensive evaluation protocol for RRSIS.
	
	\noindent\textbf{Metrics.}
	Following prior RRSIS studies, we adopt mean Intersection over Union (mIoU), global Intersection over Union (gIoU), and precision under different IoU thresholds (Pr) as evaluation metrics. For the $i$-th test sample, the IoU between the predicted mask $P_i$ and the ground-truth mask $G_i$ is defined as:
	\begin{equation}
		\mathrm{IoU}_i=\frac{|P_i \cap G_i|}{|P_i \cup G_i|}.
	\end{equation}
	The mIoU is computed by averaging the IoU values over all $N$ test samples:
	\begin{equation}
		\mathrm{mIoU}=\frac{1}{N}\sum_{i=1}^{N}\mathrm{IoU}_i.
	\end{equation}
	The gIoU is calculated by accumulating the intersection and union areas over the entire test set:
	\begin{equation}
		\mathrm{gIoU}=\frac{\sum_{i=1}^{N}|P_i \cap G_i|}{\sum_{i=1}^{N}|P_i \cup G_i|}.
	\end{equation}
	In addition, $\mathrm{Pr}$ measures the proportion of test samples whose IoU exceeds a given threshold, reflecting the reliability of predictions under different segmentation quality requirements.
	
	\subsection{Implementation Details} \label{sec:implementation_details}
	
	\noindent\textbf{Model configuration.}
	For DLG, the candidate number $K$ and Gaussian support scale $\sigma_c$ are set to 5 and 1.5 respectively. The residual guidance strength $\gamma_{\mathrm{dlg}}$ is set to 0.5 to provide moderate spatial recalibration of the fused feature, and the geometric correction weight $\lambda_g$ is set to 0.5. The foreground area threshold $\tau$ is set to 0.02 to focus guidance on potentially ambiguous cases. The response and ranking loss weights $\lambda_{\mathrm{resp}}$ and $\lambda_{\mathrm{rank}}$ are 1.0 and 1.2 respectively.  For LCR training, coarse predictions with IoU in $[0.5,0.99]$ are retained as sufficiently localized priors for contour recalibration. The residual combination coefficient $\gamma_{\mathrm{lcr}}$ is set to 1.0,and the region loss weights $\lambda_{\mathrm{ce}}$ and $\lambda_{\mathrm{dice}}$ are both set to 1.0. Notably, we use only the coarse segmenter and LCR on RefSegRS, since expressions may refer to multiple valid instances whereas DLG selects a single candidate.
	
	\noindent\textbf{Training configuration.}
	Training is performed on NVIDIA RTX 4090 GPUs using AdamW optimizer, with all input images resized to $480\times480$. The coarse model is trained with a batch size of 8 for 40 epochs at an initial learning rate of $3\times10^{-5}$ on RRSIS-D and RISBench. On RefSegRS, it is trained for 60 epochs at $5\times10^{-5}$. DLG uses 20 epochs and a batch size of 16, with initial learning rates of $8\times10^{-4}$ for the response estimator and $5\times10^{-4}$ for the ranker. LCR uses 40 epochs, a batch size of 8, and an initial learning rate of $5\times10^{-5}$. To construct offline training data, intermediate coarse checkpoints are saved at epochs 10, 15, 20, 30, and 39 on RRSIS-D and RISBench. LCR uses all five checkpoints, while DLG uses the last three and the best checkpoint to obtain features with more reliable localization. For RefSegRS, checkpoints are saved at epochs 20, 30, 40, 50, and 59 to cover its longer training schedule.

	\subsection{Accuracy Performance} \label{sec:accuracy_performance}
	We compare DiCoR with representative state-of-the-art methods from JFS paradigms. 
	
	\textbf{Results on RefSegRS.}
	As shown in Table~\ref{tab:refsegrs_comparison}, DiCoR achieves the best performance across all seven metrics among the compared methods on RefSegRS. It reaches 83.39\% gIoU and 77.58\% mIoU, exceeding MCD-Net by 2.16\% and 4.90\% respectively. The improvements are particularly pronounced at stricter IoU thresholds, where DiCoR surpasses the best-performing baseline by 12.66\% and 23.28\% respectively. These results suggest that correcting residual boundary errors contributes to more accurate mask delineation and improved segmentation performance.
	
	\begin{table*}[t]
	
	\caption{Comparison of segmentation performance on the RefSegRS test set with state-of-the-art JFS methods.}
	\label{tab:refsegrs_comparison}
	\centering
	\begin{threeparttable}
		\small
		\setlength{\tabcolsep}{1.5pt}
		\renewcommand{\arraystretch}{1.10}
		\begin{tabular}{
				>{\centering\arraybackslash}p{2.85cm}
				>{\centering\arraybackslash}p{1.65cm}
				>{\centering\arraybackslash}p{1.75cm}
				>{\centering\arraybackslash}p{1.55cm}
				|
				>{\centering\arraybackslash}p{1.13cm}
				>{\centering\arraybackslash}p{1.13cm}
				>{\centering\arraybackslash}p{1.13cm}
				>{\centering\arraybackslash}p{1.13cm}
				>{\centering\arraybackslash}p{1.13cm}
				|
				>{\centering\arraybackslash}p{1.15cm}
				>{\centering\arraybackslash}p{1.15cm}
			}
			\toprule
			\textbf{Method} & \textbf{Publication} & \textbf{Vision} & \textbf{Language}
			& \textbf{Pr@0.5} & \textbf{Pr@0.6} & \textbf{Pr@0.7} & \textbf{Pr@0.8} & \textbf{Pr@0.9}
			& \textbf{gIoU} & \textbf{mIoU} \\
			\midrule
			CMPC~\cite{huang2020referring} & CVPR'20 & R-101 & LSTM & 28.07 & 20.25 & 12.71 & 5.61 & 0.83 & 64.53 & 41.47 \\			
			BRINet~\cite{hu2020bi} & CVPR'20 & R-101 & LSTM & 22.56 & 15.74 & 9.85 & 3.52 & 0.50 & 60.16 & 32.87 \\
			CMPC+~\cite{liu2021cross}            & TPAMI'21  & R-101       & LSTM & 49.19 & 28.31 & 15.31 & 8.12 & 2.55 & 66.53 & 43.65 \\
			LAVT~\cite{yang2022lavt}             & CVPR'22   & Swin-B      & BERT & 51.84 & 30.27 & 17.34 & 9.52 & 2.09 & 71.86 & 47.40 \\
			CrossVLT~\cite{cho2023cross}         & TMM'23    & Swin-B      & BERT & 71.16 & 58.28 & 34.51 & 16.35 & 5.06 & 77.44 & 58.84 \\			
			RIS-DMMI~\cite{hu2023beyond}        & ICCV'23   & Swin-B      & BERT & 63.89 & 44.30 & 19.81 & 6.49 & 1.00 & 68.58 & 52.15 \\
			LGCE~\cite{yuan2024rrsis}           & TGRS'24   & Swin-B      & BERT & 73.75 & 61.14 & 39.46 & 16.02 & 5.45 & 76.81 & 59.96 \\
			RMSIN~\cite{liu2024rotated}         & CVPR'24   & Swin-B      & BERT & 79.20 & 65.99 & 42.98 & 16.51 & 3.25 & 75.72 & 62.58 \\
			CroBIM~\cite{dong2024cross}         & arXiv'24  & Swin-B      & BERT & 75.89 & 61.42 & 34.07 & 12.99 & 2.75 & 72.33 & 59.77 \\			
			FIANet~\cite{10816052}              & TGRS'25   & Swin-B      & BERT & 84.09 & 77.05 & 61.86 & 33.41 & 7.10 & 78.32 & 68.67 \\
			CADFormer~\cite{liu2025cadformer}   & JSTARS'25 & Swin-B      & BERT & 75.34 & 63.18 & 39.87 & 15.63 & 2.97 & 74.08 & 60.88 \\				
			Enti-TwistNet~\cite{jia2025entity}  & TGRS'25   & Swin-B      & BERT & 77.72 & 69.30 & 45.01 & 26.66 & 8.49 & 78.98 & 63.82 \\
			CroBIM-U~\cite{sun2026crobim}       & TGRS'26   & ConvNeXt-B  & BERT & 76.68 & 62.54 & 34.81 & 13.50 & 3.26 & 73.81 & 60.08 \\
			MCD-Net~\cite{zhang2026multiscale}  & TGRS'26   & Swin-B      & BERT & 85.64 & 79.75 & 69.68 & 50.41 & 14.42 & 81.23 & 72.68 \\
			S2CLNet~\cite{lei2026s2clnet}       & TGRS'26   & Swin-B      & BERT & 86.35 & 80.17 & 70.01 & 45.95 & 11.78 & 80.97 & 72.10 \\
			\midrule
			\textbf{Ours}                        & --        & Swin-B      & BERT & \textbf{87.18} & \textbf{82.11} & \textbf{75.29} & \textbf{63.07} & \textbf{37.70} & \textbf{83.39} & \textbf{77.58} \\
			\bottomrule
		\end{tabular}
	\end{threeparttable}
\end{table*}
	
	\begin{figure*}[!ht]
		\centering
		\includegraphics[width=\textwidth]{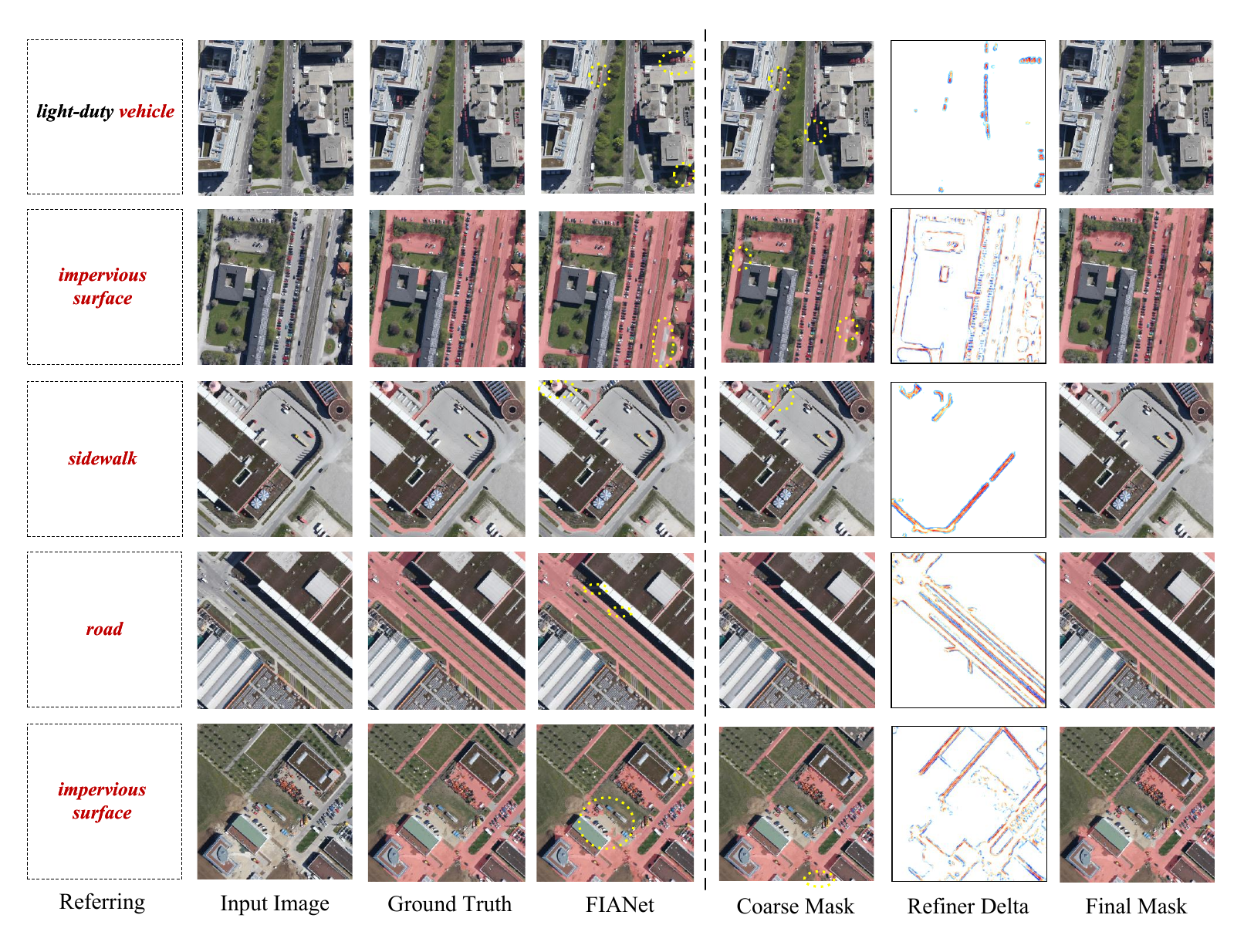}
		\caption{Qualitative results of the proposed DiCoR on the RefSegRS dataset.}
		\label{fig:refsegrs_visual}
	\end{figure*}
	
	\textbf{Results on RISBench.}
	The quantitative results on RISBench are presented in Table~\ref{tab:risbench_comparison}. DiCoR achieves the highest gIoU and mIoU among the compared methods, demonstrating its robustness under more diverse remote sensing scenes and more complex referring expressions. Although CroBIM-U achieves higher Pr@0.5 and Pr@0.6, DiCoR leads at all three stricter thresholds. The improvement is most pronounced at Pr@0.9, where DiCoR surpasses the best-performing baseline by 3.37\%. In addition, the highest mIoU suggests a benefit from improved referent discrimination, consistent with DLG’s explicit modeling of candidate competition.
	
	\begin{table*}[t]
	
	\caption{Comparison of segmentation performance on the RISBench test set with state-of-the-art JFS methods.}
	\label{tab:risbench_comparison}
	\centering
	\begin{threeparttable}
		\small
		\setlength{\tabcolsep}{1.5pt}
		\renewcommand{\arraystretch}{1.10}
		\begin{tabular}{
				>{\centering\arraybackslash}p{2.85cm}
				>{\centering\arraybackslash}p{1.65cm}
				>{\centering\arraybackslash}p{1.75cm}
				>{\centering\arraybackslash}p{1.55cm}
				|
				>{\centering\arraybackslash}p{1.13cm}
				>{\centering\arraybackslash}p{1.13cm}
				>{\centering\arraybackslash}p{1.13cm}
				>{\centering\arraybackslash}p{1.13cm}
				>{\centering\arraybackslash}p{1.13cm}
				|
				>{\centering\arraybackslash}p{1.15cm}
				>{\centering\arraybackslash}p{1.15cm}
			}
			\toprule
			\textbf{Method} & \textbf{Publication} & \textbf{Vision} & \textbf{Language}
			& \textbf{Pr@0.5} & \textbf{Pr@0.6} & \textbf{Pr@0.7} & \textbf{Pr@0.8} & \textbf{Pr@0.9}
			& \textbf{gIoU} & \textbf{mIoU} \\
			\midrule
			CMPC~\cite{huang2020referring} & CVPR'20 & R-101 & LSTM & 55.17 & 47.84 & 40.28 & 32.87 & 14.55 & 50.24 & 43.82 \\
			BRINet~\cite{hu2020bi} & CVPR'20 & R-101 & LSTM & 52.87 & 45.39 & 38.64 & 30.79 & 11.86 & 48.73 & 42.91 \\
			CMPC+~\cite{liu2021cross}            & TPAMI'21  & R-101       & LSTM & 58.02 & 49.00 & 42.53 & 35.26 & 17.88 & 53.98 & 46.73 \\
			LAVT~\cite{yang2022lavt}             & CVPR'22   & Swin-B      & BERT & 71.33 & 66.32 & 59.37 & 48.64 & 28.84 & 74.71 & 63.78 \\
			CrossVLT~\cite{cho2023cross}         & TMM'23    & Swin-B      & BERT & 70.62 & 65.05 & 57.40 & 45.80 & 26.10 & 74.33 & 62.84 \\
			RIS-DMMI~\cite{hu2023beyond}        & ICCV'23   & Swin-B      & BERT & 72.05 & 66.48 & 59.07 & 47.16 & 26.57 & 74.82 & 63.93 \\
			LGCE~\cite{yuan2024rrsis}           & TGRS'24   & Swin-B      & BERT & 71.19 & 65.92 & 58.94 & 47.87 & 27.86 & 74.34 & 63.39 \\
			RMSIN~\cite{liu2024rotated}         & CVPR'24   & Swin-B      & BERT & 74.54 & 69.51 & 62.32 & 51.53 & 30.58 & 73.35 & 66.86 \\
			CroBIM~\cite{dong2024cross}         & arXiv'24  & Swin-B      & BERT & 75.75 & 70.34 & 63.12 & 51.12 & 28.45 & 73.61 & 67.32 \\
			FIANet~\cite{10816052}              & TGRS'25   & Swin-B      & BERT & 74.80 & 69.86 & 63.16 & 52.70 & 31.87 & 74.23 & 66.83 \\
			CADFormer~\cite{liu2025cadformer}   & JSTARS'25 & Swin-B      & BERT & 76.27 & 71.35 & 64.21 & 52.91 & 31.59 & 74.34 & 67.80 \\			
			LSCF~\cite{ma2025lscf}              & TGRS'25   & Swin-B      & BERT & 76.08 & 71.29 & 64.96 & 55.13 & 36.73 & 74.88 & 68.53 \\			
			Enti-TwistNet~\cite{jia2025entity}  & TGRS'25   & Swin-B      & BERT & 76.13 & 71.47 & 63.48 & 52.38 & 27.46 & 74.07 & 67.50 \\
			CroBIM-U~\cite{sun2026crobim}       & TGRS'26   & ConvNeXt-B  & BERT & \textbf{77.73} & \textbf{73.40} & 67.05 & 56.85 & 35.52 & 73.04 & 69.62 \\
			S2CLNet~\cite{lei2026s2clnet}       & TGRS'26   & Swin-B      & BERT & 76.79 & 71.91 & 65.18 & 54.77 & 34.51 & 75.28 & 68.54 \\
			\midrule
			\textbf{Ours}                        & --        & Swin-B      & BERT & 77.28 & 72.83 & \textbf{67.38} & \textbf{58.93} & \textbf{40.10} & \textbf{75.53} & \textbf{70.10} \\
			\bottomrule
		\end{tabular}
	\end{threeparttable}
\end{table*}
	
	\begin{figure*}[!ht]
		\captionsetup{belowskip=-10pt} 
		\centering
		\includegraphics[width=\textwidth]{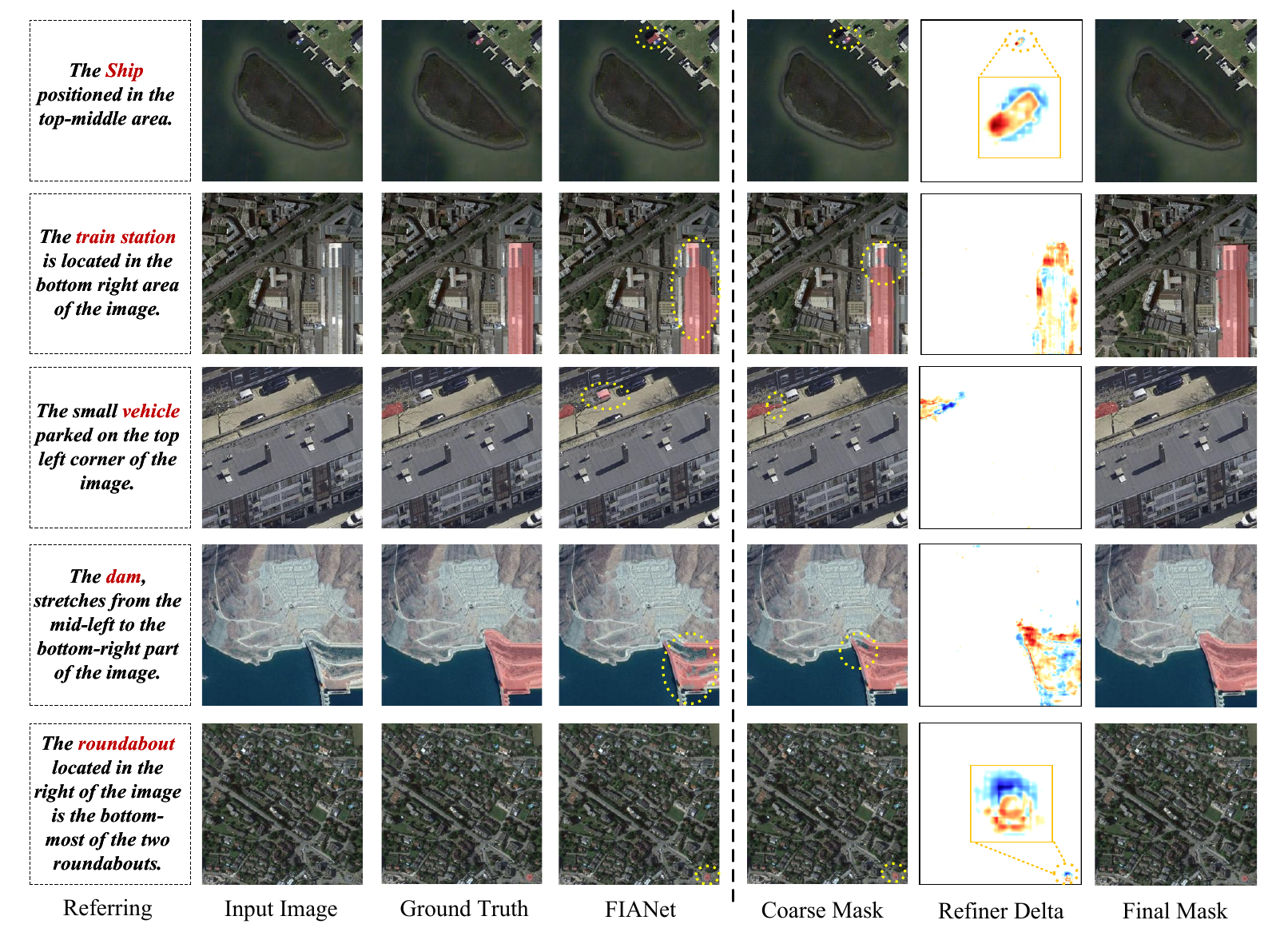}
		\caption{Qualitative results of the proposed DiCoR on the RISBench dataset.}
		\label{fig:risbench_visual}
	\end{figure*}
	
	\textbf{Results on RRSIS-D.}
	As reported in Table~\ref{tab:rrsisd_comparison}, DiCoR achieves the highest mIoU and precision at all five IoU thresholds among the compared methods on RRSIS-D. The largest precision gain occurs at Pr@0.6, where DiCoR achieves 70.11\%, surpassing S2CLNet by 1.22\%, demonstrating its superior performance in accurately identifying the target objects. Although its gIoU trails MCD-Net by 0.37\%, the higher mIoU and precision support its effectiveness in segmenting individual referred targets.

	The class-wise results in Table~\ref{tab:rrsisd_classwise_comparison} further demonstrate that DiCoR achieves improvements across diverse categories. The gains on “vehicle” and “bridge” suggest that DLG helps distinguish the referred target from similar alternatives, providing reliable localization for subsequent contour refinement. LCR then corrects local boundary deviations, particularly beneficial for small targets in remote sensing imagery. These benefits are less pronounced for “chimney” and “harbor”, where relatively large targets with fewer ambiguous alternatives may leave less scope for DLG. In addition, the elongated structures and long contours of harbors can involve substantial missing regions beyond the local correction range of LCR, making complete recovery more challenging.

	\begin{table*}[t]
	
	\caption{Comparison of segmentation performance on the RRSIS-D test set with state-of-the-art JFS methods.}
	\label{tab:rrsisd_comparison}
	\centering
	\begin{threeparttable}
		\small
		\setlength{\tabcolsep}{1.5pt}
		\renewcommand{\arraystretch}{1.10}
		\begin{tabular}{
				>{\centering\arraybackslash}p{2.85cm}
				>{\centering\arraybackslash}p{1.65cm}
				>{\centering\arraybackslash}p{1.75cm}
				>{\centering\arraybackslash}p{1.55cm}
				|
				>{\centering\arraybackslash}p{1.13cm}
				>{\centering\arraybackslash}p{1.13cm}
				>{\centering\arraybackslash}p{1.13cm}
				>{\centering\arraybackslash}p{1.13cm}
				>{\centering\arraybackslash}p{1.13cm}
				|
				>{\centering\arraybackslash}p{1.15cm}
				>{\centering\arraybackslash}p{1.15cm}
			}
			\toprule
			\textbf{Method} & \textbf{Publication} & \textbf{Vision} & \textbf{Language}
			& \textbf{Pr@0.5} & \textbf{Pr@0.6} & \textbf{Pr@0.7} & \textbf{Pr@0.8} & \textbf{Pr@0.9}
			& \textbf{gIoU} & \textbf{mIoU} \\
			\midrule
			CMPC~\cite{huang2020referring} & CVPR'20 & R-101 & LSTM & 55.83 & 47.40 & 36.94 & 25.45 & 9.19 & 69.22 & 49.24 \\
			BRINet~\cite{hu2020bi} & CVPR'20 & R-101 & LSTM & 56.90 & 48.77 & 39.12 & 27.03 & 8.73 & 69.88 & 49.65 \\
			CMPC+~\cite{liu2021cross}            & TPAMI'21  & R-101       & LSTM & 57.65 & 47.51 & 36.97 & 24.33 & 7.78 & 68.64 & 50.24 \\
			LAVT~\cite{yang2022lavt}             & CVPR'22   & Swin-B      & BERT & 69.52 & 63.63 & 53.29 & 41.60 & 24.94 & 77.19 & 61.04 \\
			CrossVLT~\cite{cho2023cross}         & TMM'23    & Swin-B      & BERT & 70.38 & 63.83 & 52.86 & 42.11 & 25.02 & 76.32 & 61.00 \\
			RIS-DMMI~\cite{hu2023beyond}        & ICCV'23   & Swin-B      & BERT & 68.74 & 60.96 & 50.33 & 38.38 & 21.63 & 76.20 & 60.12 \\			
			LGCE~\cite{yuan2024rrsis}           & TGRS'24   & Swin-B      & BERT & 67.65 & 61.53 & 51.45 & 39.62 & 23.33 & 76.34 & 59.37 \\
			RMSIN~\cite{liu2024rotated}         & CVPR'24   & Swin-B      & BERT & 74.26 & 67.25 & 55.93 & 42.55 & 24.53 & 77.79 & 64.20 \\
			CroBIM~\cite{dong2024cross}         & arXiv'24  & Swin-B      & BERT & 74.58 & 67.57 & 55.59 & 41.63 & 23.56 & 75.99 & 64.46 \\
			FIANet~\cite{10816052}              & TGRS'25   & Swin-B      & BERT & 74.46 & 66.96 & 56.31 & 42.83 & 24.13 & 76.91 & 64.01 \\
			CADFormer~\cite{liu2025cadformer}   & JSTARS'25 & Swin-B      & BERT & 74.20 & 67.62 & 55.59 & 42.37 & 23.59 & 77.26 & 63.77 \\
			LSCF~\cite{ma2025lscf}              & TGRS'25   & Swin-B      & BERT & 74.30 & 67.69 & 56.32 & 43.08 & 25.67 & 77.42 & 64.25 \\
			MCD-Net~\cite{zhang2026multiscale}  & TGRS'26   & Swin-B      & BERT & 75.58 & 68.74 & 57.51 & 44.15 & 25.97 & \textbf{78.14} & 65.05 \\
			CroBIM-U~\cite{sun2026crobim}       & TGRS'26   & ConvNeXt-B  & BERT & 75.60 & 67.68 & 56.47 & 42.57 & 24.16 & 76.70 & 65.07 \\
			S2CLNet~\cite{lei2026s2clnet}       & TGRS'26   & Swin-B      & BERT & 74.89 & 68.89 & 56.68 & 44.13 & 25.60 & 77.69 & 64.49 \\
			\midrule
			\textbf{Ours}                        & --        & Swin-B      & BERT & \textbf{75.86} & \textbf{70.11} & \textbf{58.33} & \textbf{44.22} & \textbf{26.26} & 77.77 & \textbf{65.34} \\
			\bottomrule
		\end{tabular}
	\end{threeparttable}
	\end{table*}

	\begin{table}[!t]
		
		\centering
		\captionsetup{font={footnotesize,},skip=4pt}
		\caption{The results on each category on RRSIS-D dataset.}
		\label{tab:rrsisd_classwise_comparison}
		\footnotesize
		\setlength{\tabcolsep}{3pt}
		\renewcommand{\arraystretch}{1.06}
		\begin{tabular}{
		>{\centering\arraybackslash}p{3.2cm}
		>{\centering\arraybackslash}p{1.05cm}
		>{\centering\arraybackslash}p{1.05cm}
		>{\centering\arraybackslash}p{1.05cm}
		>{\centering\arraybackslash}p{1.05cm}
		}
		\toprule
	Category & LGCE & RMSIN & FIANet & Ours \\
	\midrule
	Airplane & 68.11 & 68.08 & 68.66 & \textbf{69.58} \\
	Airport & 56.43 & 56.11 & 57.07 & \textbf{58.93} \\
	Golf field & 77.19 & 76.68 & 77.35 & \textbf{78.02} \\
	Expressway service area & 70.93 & 66.93 & 70.44 & \textbf{73.56} \\
	Baseball field & 84.90 & 83.09 & 84.87 & \textbf{85.97} \\
	Stadium & 82.54 & 81.91 & 82.00 & \textbf{84.37} \\
	Ground track field & 73.33 & 73.65 & 76.99 & \textbf{77.66} \\
	Storage tank & 74.37 & 72.26 & 74.86 & \textbf{77.18} \\
	Basketball court & 68.44 & 68.42 & 68.41 & \textbf{70.72} \\
	Chimney & 75.63 & 76.68 & \textbf{78.48} & 77.47 \\
	Tennis court & 67.67 & 70.14 & 70.01 & \textbf{72.91} \\
	Overpass & 58.19 & \textbf{62.67} & 61.30 & 61.27 \\
	Train station & 63.48 & 64.64 & \textbf{65.96} & 65.54 \\
	Ship & 61.63 & 65.71 & 64.82 & \textbf{68.48} \\
	Expressway toll station & 64.54 & 68.70 & 71.31 & \textbf{73.10} \\
	Dam & 60.47 & 60.40 & \textbf{62.03} & 61.36 \\
	Harbor & 34.24 & 36.74 & \textbf{37.94} & 35.95 \\
	Bridge & 43.12 & 47.63 & 49.66 & \textbf{51.03} \\
	Vehicle & 40.76 & 41.99 & 46.72 & \textbf{49.09} \\
	Windmill & 56.51 & 60.18 & 60.25 & \textbf{60.44} \\
	\midrule
	Average & 64.12 & 65.13 & 66.46 & \textbf{67.63} \\
	\bottomrule
		\end{tabular}
	\end{table}
	
	\begin{figure*}[!ht]
		\centering
		\includegraphics[width=\textwidth]{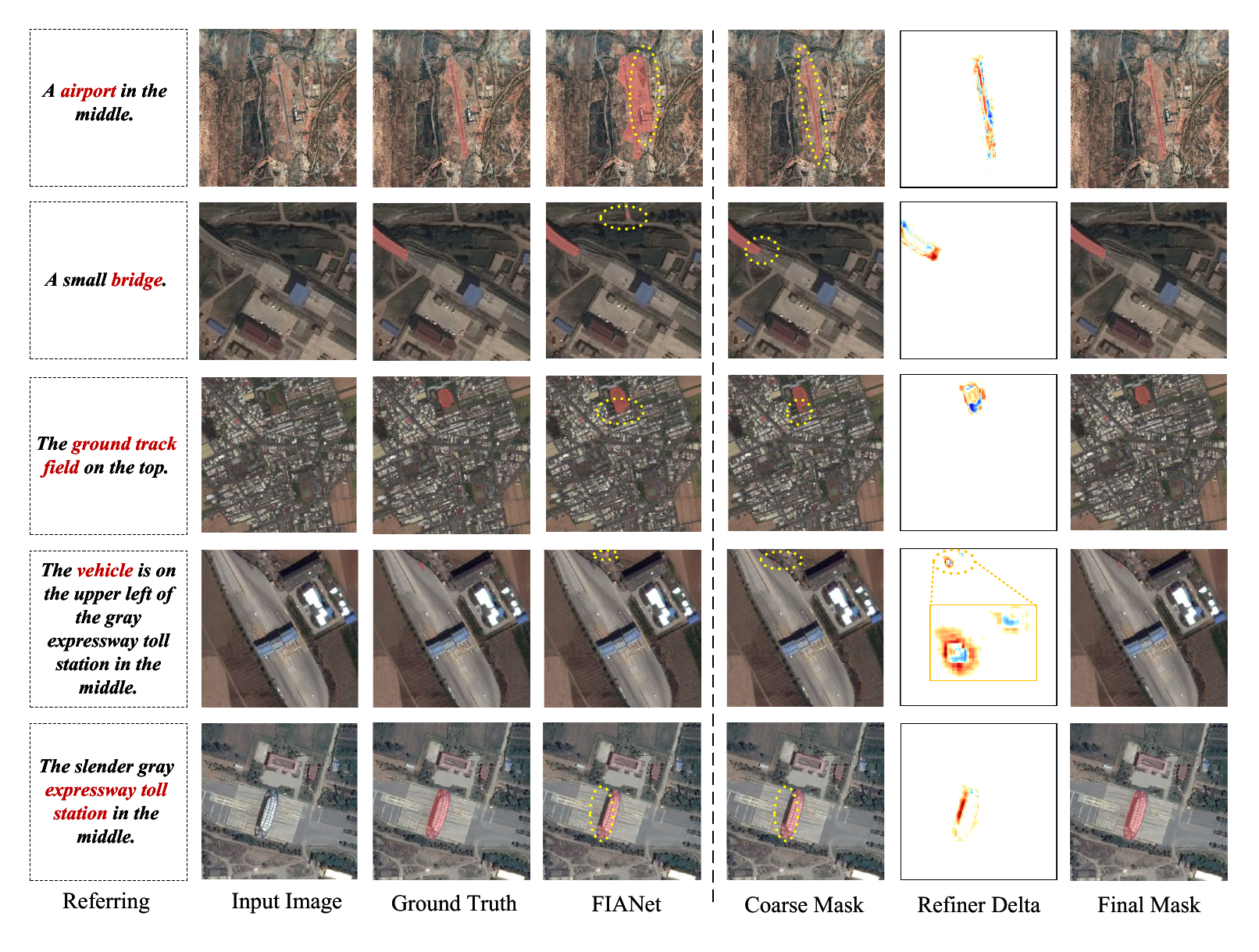}
		\caption{Qualitative results of the proposed DiCoR on the RRSIS-D dataset.}
		\label{fig:rrsisd_visual}
	\end{figure*}
	
	\textbf{Visualization analysis.}
	Figs.~\ref{fig:refsegrs_visual}, \ref{fig:risbench_visual}, and \ref{fig:rrsisd_visual} present qualitative results on RefSegRS, RISBench, and RRSIS-D respectively. 
	On RefSegRS, the referring expressions are relatively simple, whereas the referred targets often occupy large and spatially extended regions, such as roads, impervious surfaces, and sidewalks. In these cases, DiCoR produces more complete masks and more coherent contours than both FIANet and the coarse prediction, indicating that the recalibration process can improve structural consistency while preserving reliable coarse localization. On RISBench and RRSIS-D, the scenes are more complex and the expressions usually involve more detailed spatial relations and semantic cues, making both target discrimination and boundary delineation more challenging. DiCoR shows clearer advantages in focusing on the target referent and suppressing irrelevant responses from surrounding regions. Notably, the $\Delta Z$ heatmaps are concentrated around uncertain regions, highlighting the localized residual correction performed by LCR. This contour-aware recalibration is particularly beneficial for small remote sensing objects, such as vehicles and ships, where slight boundary deviations can substantially degrade high-threshold precision.
	
	\subsection{Efficiency Performance}\label{sec:efficiency_performance}
	To further evaluate the accuracy--efficiency trade-off of DiCoR, Table~\ref{tab:efficiency_comparison} compares DiCoR with representative JFS and DPS methods on RISBench. Following the weighted-product formulation of MnasNet~\cite{tan2019mnasnet}, we define an accuracy--efficiency index (AEI) that combines segmentation accuracy with inference throughput:
	\begin{equation}
	    \mathrm{AEI}_i
	    =
	    \mathrm{mIoU}_i
	    \left(
	        \frac{\mathrm{FPS}_i}
	        {\max_{k \in \mathcal{M}}\mathrm{FPS}_k}
	    \right)^{\gamma},
	\end{equation}
	where $\mathcal{M}$ denotes the set of compared methods and $\gamma$ is set to $0.07$, matching the exponent magnitude in MnasNet~\cite{tan2019mnasnet}. A larger AEI indicates a more favorable overall trade-off.
	
	Compared with existing JFS methods, the main advantage of DiCoR lies in its stronger segmentation accuracy, with the highest mIoU among the evaluated JFS methods. Although DiCoR has a larger model size and higher computational cost than some JFS baselines, DLG and LCR contain 1.63M and 31.04M parameters respectively, and their simple computational structures support relatively fast inference. Under the adopted weighting, DiCoR also achieves the highest AEI within the JFS group.
		
	Compared with DPS methods, DiCoR exhibits a clear efficiency advantage. DPS methods typically rely on large vision-language encoders or foundation segmenters, which introduce considerable computational overhead and slow down inference. In contrast, DiCoR retains an efficient JFS inference pipeline and incorporates lightweight auxiliary modules. As a result, DiCoR achieves the highest AEI among all compared methods, demonstrating a favorable balance between segmentation accuracy and inference efficiency.
	
	\begin{table*}[t]
		\caption{Efficiency comparison with representative JFS and DPS methods on RISBench dataset.}
		\label{tab:efficiency_comparison}
		\centering
		\begin{threeparttable}
			\footnotesize
			\setlength{\tabcolsep}{4pt}
			\renewcommand{\arraystretch}{1.05}
			\begin{tabular}{
					>{\centering\arraybackslash}p{1.20cm}
					>{\centering\arraybackslash}p{2.40cm}
					>{\centering\arraybackslash}p{3cm}
					>{\centering\arraybackslash}p{2.15cm}
					>{\centering\arraybackslash}p{1.5cm}
					>{\centering\arraybackslash}p{1.5cm}
					>{\centering\arraybackslash}p{1.25cm}
					>{\centering\arraybackslash}p{1.25cm}
				}
				\toprule
				\textbf{Paradigm} & \textbf{Method} & \textbf{Backbone} & \textbf{Segmenter} & \textbf{Params (M)} & \textbf{GFLOPs} & \textbf{FPS} & \textbf{AEI} \\
				\midrule
				\multirow{4}{*}{JFS}
				& LAVT~\cite{yang2022lavt}           & Swin-B + BERT & Conv. decoder & \textbf{227.74} & \textbf{198.82} & \textbf{28.65} & 63.78 \\
				& RMSIN~\cite{liu2024rotated}        & Swin-B + BERT & Conv. decoder & 240.04 & 200.33 & 17.50 & 64.59 \\
				& FIANet~\cite{10816052}             & Swin-B + BERT & Conv. decoder & 256.17 & 210.44 & 24.18 & 66.04 \\
				& MCD-Net~\cite{zhang2026multiscale} & Swin-B + BERT & Conv. decoder & 354.97 & 482.73 & 7.68 & 62.79 \\
				\midrule
				\multirow{4}{*}{DPS}
				& RSRefSeg-1~\cite{chen2025rsrefseg}  & CLIP & SAM & 984.28 & 1610.75 & 6.62 & 62.95 \\
				& RSRefSeg-2~\cite{chen2025rsrefseg2} & CLIP & SAM & 1447.13 & 2563.51 & 5.40 & 64.57 \\
				& SegEarth-R1~\cite{li2025segearth}  & Swin-B + Phi-1.5B & Mask2Former & 1589.29 & 2750.90 & 5.27 & 61.19 \\
				& SegEarth-R2~\cite{xin2025segearth} & Swin-B/SigLIP + Phi-2B & Mask2Former & 3325.06 & 6018.13 & 2.41 & 59.28 \\
				\midrule
				\textbf{JFS}
				& \textbf{Ours} & Swin-B + BERT & Conv. decoder & 274.76 & 443.25 & 15.60 & \textbf{67.18} \\
				\bottomrule
			\end{tabular}
			\begin{tablenotes}[flushleft]
				\footnotesize
				\item All methods are evaluated based on their publicly available implementations on a NVIDIA RTX 4090 GPU. The image input resolution and maximum text length are kept identical for fair comparison.
			\end{tablenotes}
		\end{threeparttable}
	\end{table*}
	
	Notably, SAM3-assisted distractor annotation for DLG training introduces an additional offline preprocessing cost. On a single NVIDIA RTX 4090 GPU, preprocessing the RISBench training set requires an average of 233.1 ms per sample, corresponding to a total processing time of 1.71 h and a peak GPU memory usage of 4.21 GB. This overhead is acceptable for offline data preparation, as the training set requires only a single preprocessing pass, after which the generated distractor annotations are cached and reused throughout DLG training. SAM3 is therefore not invoked during subsequent training epochs or deployment, allowing DiCoR to benefit from auxiliary distractor supervision while preserving its inference efficiency.

	\subsection{Ablation Study} \label{sec:ablation_study}
	We conduct ablation experiments on the RISBench dataset to examine the contributions of the proposed components. To provide a clear and intuitive analysis of how each module affects performance, all ablation variants are constructed by modifying the same baseline model with a single component replaced at a time. Unless otherwise specified, all experiments are performed under the same training protocol.
	
	\textbf{1) Overall component ablation.} Table~\ref{tab:main_ablation} demonstrates the effectiveness of DLG and LCR. Their combination achieves the best results across all four metrics, increasing mIoU from 67.98\% to 70.10\% and Pr@0.9 from 32.10\% to 40.10\%. Although DLG alone improves overall mIoU and gIoU by 0.32\% and 0.05\% respectively, the limited gIoU gain is expected given its primary benefits on smaller targets. Corrections to these targets contribute relatively few pixels to the global metric, while resolving referent selection errors provides more reliable target support for LCR, helping the combined model achieve higher accuracy. Since target scales vary greatly in remote sensing imagery, we further assess performance across scales to identify where each module contributes most.
	
	Specifically, we measure target scale using the ground-truth foreground area ratio $\rho$. With thresholds informed by the corresponding distribution in coarse masks, we divide the test samples into four groups, including tiny ($\rho\leq0.1\%$), small ($0.1\%<\rho\leq0.5\%$), medium ($0.5\%<\rho\leq2\%$), and large ($\rho>2\%$). DLG primarily benefits tiny, small, and medium targets, increasing mIoU by 0.82\%, 0.54\%, and 0.34\% respectively. These gains are consistent with resolving competition among small referents and visually similar distractors. DLG provides limited improvement on large targets because its activation is constrained by a foreground area threshold. LCR improves mIoU across all four scales, while increasing Pr@0.9 by 10.04\% and 11.16\% on tiny and small targets respectively. Combining the modules yields the highest mIoU for tiny, small, and medium targets, while performing nearly identically to LCR alone on large targets.

	\begin{table}[htbp]
		\captionsetup{justification=centering}
		\caption{Component ablation on RISBench with overall results and a breakdown by target scale.}
		\label{tab:main_ablation}
		\centering
		\footnotesize
		
		\renewcommand{\arraystretch}{1.10}
		\setlength{\tabcolsep}{2.5pt}
		\begin{tabular}{
			>{\centering\arraybackslash}p{1.15cm}
			>{\centering\arraybackslash}p{0.65cm}
			>{\centering\arraybackslash}p{0.65cm}
			>{\centering\arraybackslash}p{1.15cm}
			>{\centering\arraybackslash}p{1.15cm}
			>{\centering\arraybackslash}p{1.15cm}
			>{\centering\arraybackslash}p{1.15cm}}
			\toprule
			\multirow{2}{*}{\textbf{Scale}} & \multicolumn{2}{c}{\textbf{Modules}} & \multicolumn{4}{c}{\textbf{Performance (\%)}} \\
			\cmidrule(lr){2-3} \cmidrule(lr){4-7}
			& \textbf{DLG} & \textbf{LCR} & \textbf{Pr@0.5} & \textbf{Pr@0.9} & \textbf{mIoU} & \textbf{gIoU} \\
			\midrule
			\multicolumn{7}{c}{\textit{Overall performance}} \\
			\addlinespace[2pt]
			\multirow{4}{*}{\textbf{Total}} & -- & -- & 76.25 & 32.10 & 67.98 & 74.71 \\
			 & $\checkmark$ & -- & 76.67 & 32.16 & 68.30 & 74.76 \\
			 & -- & $\checkmark$ & 76.76 & 39.84 & 69.75 & 75.47 \\
			 & $\checkmark$ & $\checkmark$ & \textbf{77.28} & \textbf{40.10} & \textbf{70.10} & \textbf{75.53} \\
			\midrule
			\multicolumn{7}{c}{\textit{Performance across target scales}} \\
			\addlinespace[2pt]
			\multirow{4}{*}{Tiny} & -- & -- & 54.21 & 7.55 & 47.68 & 28.13 \\
			 & $\checkmark$ & -- & 55.31 & 7.44 & 48.50 & \textbf{28.95} \\
			 & -- & $\checkmark$ & 54.33 & 17.59 & 50.32 & 28.28 \\
			 & $\checkmark$ & $\checkmark$ & \textbf{55.88} & \textbf{18.28} & \textbf{51.11} & 28.92 \\
			\cmidrule(lr){1-7}
			\multirow{4}{*}{Small} & -- & -- & 70.21 & 22.00 & 61.90 & 43.77 \\
			 & $\checkmark$ & -- & 70.91 & 22.12 & 62.44 & 44.41 \\
			 & -- & $\checkmark$ & 71.13 & 33.16 & 64.26 & 44.34 \\
			 & $\checkmark$ & $\checkmark$ & \textbf{71.78} & \textbf{33.60} & \textbf{64.86} & \textbf{45.02} \\
			\cmidrule(lr){1-7}
			\multirow{4}{*}{Medium} & -- & -- & 76.50 & 29.80 & 67.80 & 57.98 \\
			 & $\checkmark$ & -- & 76.89 & 29.94 & 68.14 & 58.30 \\
			 & -- & $\checkmark$ & 77.03 & 37.90 & 69.58 & 58.77 \\
			 & $\checkmark$ & $\checkmark$ & \textbf{77.73} & \textbf{38.10} & \textbf{69.98} & \textbf{59.18} \\
			\cmidrule(lr){1-7}
			\multirow{4}{*}{Large} & -- & -- & 87.68 & 49.23 & 79.18 & 77.16 \\
			 & $\checkmark$ & -- & 87.68 & 49.23 & 79.16 & 77.16 \\
			 & -- & $\checkmark$ & \textbf{87.95} & 53.18 & \textbf{80.20} & \textbf{77.97} \\
			 & $\checkmark$ & $\checkmark$ & 87.93 & \textbf{53.20} & 80.19 & \textbf{77.97} \\
			\bottomrule
		\end{tabular}
	\end{table}

	\textbf{2) Ablation of DLG design and supervision.} We examine the effectiveness of DLG’s ranking design and distractor supervision, as shown in Table~\ref{tab:dlg_design_ablation}. For candidate selection, directly choosing the candidate with the strongest response (Top response) produces the weakest results, indicating that response magnitude alone is insufficient for reliable referent discrimination. The visual-only ranker (Visual-Only) improves performance by explicitly comparing candidate visual features and enhancing discrimination among competing regions. The visual-text ranker (Visual-Text) further reweights textual representations according to candidate features, providing candidate-specific linguistic evidence for more effective cross-modal ranking and achieving the best results across all four metrics. For distractor supervision, all evaluated sources improve mIoU and gIoU over the setting without distractor annotations, while differences among sources remain modest. Since Grounded-SAM and Grounded-SAM 2 rely on GroundingDINO for object localization, the quality of their distractor annotations is constrained by its detection accuracy.

	\begin{table}[htbp]
	    \captionsetup{justification=centering}
	    \caption{Ablation of candidate ranking and distractor supervision in DLG on RISBench dataset. ``--'' denotes no distractor supervision.}
	    \label{tab:dlg_design_ablation}
	    \centering
	    
	    \begin{threeparttable}
	        \footnotesize
	        \renewcommand{\arraystretch}{1.10}
	        \setlength{\tabcolsep}{0.75pt}
	        \begin{tabular}{
	            >{\centering\arraybackslash}p{1.20cm}
	            >{\centering\arraybackslash}p{2.90cm}
	            >{\centering\arraybackslash}p{1.1cm}
	            >{\centering\arraybackslash}p{1.1cm}
	            >{\centering\arraybackslash}p{1.1cm}
	            >{\centering\arraybackslash}p{1.1cm}}
	            \toprule
	            \multirow{2}{*}{\textbf{Ablation}} &
	            \multirow{2}{*}{\textbf{Setting}} &
	            \multicolumn{4}{c}{\textbf{Performance (\%)}} \\
	            \cmidrule(lr){3-6}
	            & & \textbf{Pr@0.5} & \textbf{Pr@0.9} & \textbf{mIoU} & \textbf{gIoU} \\
	            \midrule
	            \multirow{3}{*}{\makecell{Ranking\\strategy}}
	            & Top response & 76.14 & 39.85 & 69.27 & 75.32 \\
	            & Visual-Only       & 77.00 & 39.86 & 69.84 & 75.39 \\
	            & Visual-Text       & \textbf{77.28} & \textbf{40.10} & \textbf{70.10} & \textbf{75.53} \\
	            \midrule
	            \multirow{6}{*}{\makecell{Distractor\\source}}
	            & --                & 76.64 & 39.90 & 69.56 & 75.44 \\
	            & SAM~\cite{kirillov2023segment} & 77.14 & 40.06 & 70.01 & 75.47 \\
	            & SAM 2~\cite{ravi2025sam2} & 77.11 & \textbf{40.11} & 70.03 & 75.50 \\
	            & Grounded-SAM~\cite{ren2024grounded} & 76.85 & 39.97 & 69.78 & 75.47 \\
	            & Grounded-SAM 2~\cite{ren2024groundedsam2} & 77.02 & 39.89 & 69.92 & 75.45 \\
	            & SAM3~\cite{carion2025sam} & \textbf{77.28} & 40.10 & \textbf{70.10} & \textbf{75.53} \\
	            \bottomrule
	        \end{tabular}
	    \end{threeparttable}
	\end{table}

	\textbf{3) Ablation of DLG localization mechanism.} We further investigate how candidate generation and selection contribute to DLG’s localization corrections, as shown in Table~\ref{tab:dlg_candidate_diagnostics}. Following the activation rule, DLG is applied to 10,290 test samples whose coarse-mask foreground area ratio falls below $\tau$. Notably, Large samples are included when their coarse-mask foreground ratio falls below $\tau$ despite a ground-truth ratio above it. Samples with baseline coarse IoU $\geq 0.5$ are classified as Easy subset, and those with IoU $< 0.5$ as Hard subset. Recall@5 measures whether the five candidates include the target, while Rank@1 measures whether the top-ranked candidate matches the target. A candidate is considered matched if its peak lies within the ground-truth mask at the response map resolution. 
	
\begin{figure}[htb]
		\vspace{-4pt}
		\captionsetup{belowskip=-2pt}
		\centering
		\includegraphics[width=\linewidth]{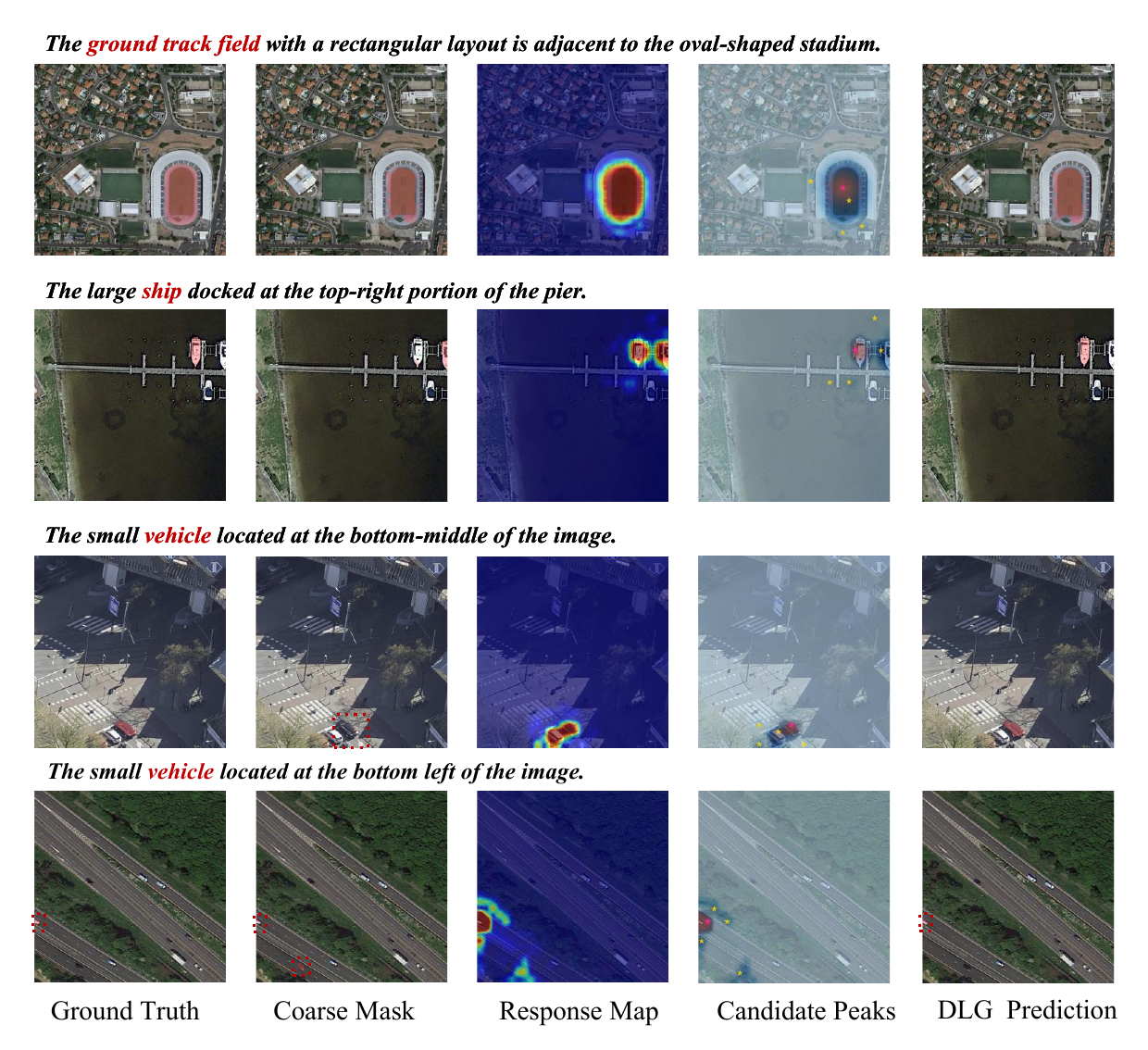}
		\caption{Visualization of DLG mechanism across target scales. From top to bottom, the rows show samples with large, medium, small, and tiny targets respectively. Red stars mark the selected candidate peaks, while yellow stars indicate the remaining candidate peaks.}
		\label{fig:dlg_visualization}
	\end{figure}

	According to the results, the easy subset achieve higher Recall@5 and Rank@1, but correct candidate selection does not necessarily translate into improved mask accuracy. Misalignment between the injected support and the target extent may introduce contour deviations and slight mIoU losses, especially for smaller targets. These residual contour discrepancies highlight the complementary role of LCR in refining target boundaries beyond DLG’s localization guidance. For hard samples, DLG can correct initial localization errors by selecting a retained target-matched candidate and injecting its spatial support. This is reflected in mIoU gains of 1.54\%–1.93\% for hard samples with tiny, small, or medium targets. Fig.~\ref{fig:dlg_visualization} further illustrates this behavior. Large targets often exhibit limited referent ambiguity, as illustrated by the first sample, where responses are concentrated within the target and DLG largely preserves its localization. In the remaining examples, target responses coexist with distractor responses despite incorrect coarse predictions, and DLG redirects segmentation toward the intended instance through candidate ranking and spatial guidance.

	\begin{table}[htbp]
	    \captionsetup{justification=centering}
	    \caption{Candidate-level performance of DLG across difficulty and target scales on RISBench test set.}
	    \label{tab:dlg_candidate_diagnostics}
	    \centering
	    
	    \begin{threeparttable}
	        \footnotesize
	        \renewcommand{\arraystretch}{1.10}
	        \setlength{\tabcolsep}{2pt}
	        \begin{tabular}{
	            >{\centering\arraybackslash}p{0.95cm}
	            >{\centering\arraybackslash}p{1.20cm}
	            >{\centering\arraybackslash}p{1.10cm}
	            >{\centering\arraybackslash}p{1.55cm}
	            >{\centering\arraybackslash}p{1.35cm}
	            >{\centering\arraybackslash}p{1.40cm}}
	            \toprule
	            \multirow{2}{*}{\textbf{Subset}} &
	            \multirow{2}{*}{\textbf{Scale}} &
	            \multirow{2}{*}{\textbf{Samples}} &
	            \multicolumn{2}{c}{\textbf{Candidate metrics (\%)}} &
	            \multirow{2}{*}{\makecell{\textbf{$\Delta$mIoU}\\\textbf{(pp)}}} \\
	            \cmidrule(lr){4-5}
	            & & & \textbf{Recall@5} & \textbf{Rank@1} & \\
	            \midrule
	            \multirow{4}{*}{Easy}
	            & Tiny   & 940   & 85.00 & 84.15 & $-0.03$ \\
	            & Small  & 3,485 & 96.01 & 94.92 & $+0.12$ \\
	            & Medium & 2,634 & 97.99 & 95.63 & $-0.05$ \\
	            & Large  & 99    & 98.99 & 93.94 & $-0.17$ \\
	            \midrule
	            \multirow{4}{*}{Hard}
	            & Tiny   & 788   & 73.10 & 66.37 & $+1.84$ \\
	            & Small  & 1,440 & 87.85 & 79.44 & $+1.54$ \\
	            & Medium & 700   & 91.86 & 83.57 & $+1.93$ \\
	            & Large  & 204   & 87.75 & 58.33 & $-0.37$ \\
	            \bottomrule
	        \end{tabular}
	    \end{threeparttable}
	\end{table}
	
\begin{figure}[htb]
		\vspace{-4pt}
		\captionsetup{belowskip=-2pt}
		\centering
		\includegraphics[width=\linewidth]{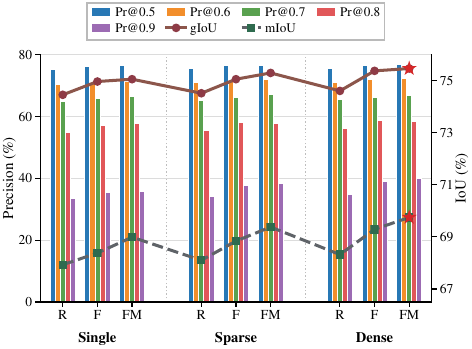}
		\caption{Ablation of LCR training data construction. Single denotes coarse predictions from the final checkpoint, whereas Sparse and Dense denote predictions sampled at increasing densities from multiple training checkpoints. R, F, and FM denote raw predictions, predictions filtered by localization quality, and filtered predictions with morphological perturbation respectively.}
		\label{fig:lcr_prior_ablation}
	\end{figure}

	\textbf{4) Ablation of LCR training strategy.} Figure~\ref{fig:lcr_prior_ablation} examines the effects of checkpoint sampling, localization filtering, and morphological perturbation on LCR pretraining. Priors from the final checkpoint alone provide limited error diversity and yield modest recalibration performance. In contrast, sparse and dense sampling across training stages progressively increase the diversity of contour deviations and prediction quality. Filtering improves all checkpoint sampling settings, with particularly pronounced gains in Pr@0.9, suggesting that excluding poorly localized priors benefits residual recalibration. Morphological perturbation introduces additional boundary variations through local expansion and shrinkage, achieving the strongest overall performance under dense checkpoint sampling.
	
	\textbf{5) Ablation of LCR calibration mechanism.} To examine LCR’s dependence on initial mask quality, we group RISBench test samples into five intervals according to baseline coarse-mask IoU, keeping both the baseline and refiner fixed. As shown in Table~\ref{tab:lcr_calibration_mechanism}, samples in $[0,0.3)$ exhibit limited recovery, with an mIoU gain of 0.63\% and improvements in only 20.72\% of samples. Larger gains of 3.81\% and 2.94\% points are observed in $[0.5,0.7)$ and $[0.7,0.9)$ respectively, consistent with more effective contour correction when the coarse mask provides reliable target support. For already accurate masks in $[0.9,1.0]$, 69.98\% of samples improve, although the smaller remaining error limits the gain to 0.51\%. Together, these results suggest that LCR’s refinement benefits depend on both the reliability of the initial target support and the remaining scope for contour correction.

	\begin{table}[htbp]
		\captionsetup{justification=centering}
		\caption{LCR refinement gains across coarse IoU intervals on the RISBench dataset.}
		\label{tab:lcr_calibration_mechanism}
		\centering
		\footnotesize
		
		\renewcommand{\arraystretch}{1.10}
		\setlength{\tabcolsep}{2.5pt}
		\begin{tabular}{
			>{\centering\arraybackslash}p{1.25cm}
			>{\centering\arraybackslash}p{1.10cm}
			>{\centering\arraybackslash}p{1.30cm}
			>{\centering\arraybackslash}p{1.30cm}
			>{\centering\arraybackslash}p{0.95cm}
			>{\centering\arraybackslash}p{1.60cm}}
			\toprule
			\makecell{\textbf{Coarse}\\\textbf{IoU bin}} &
			\textbf{Samples} &
			\makecell{\textbf{Coarse}\\\textbf{mIoU (\%)}} &
			\makecell{\textbf{Refined}\\\textbf{mIoU (\%)}} &
			\makecell{\textbf{$\Delta$mIoU}\\\textbf{(pp)}} &
			\makecell{\textbf{Improved}\\\textbf{samples (\%)}} \\
			\midrule
			$[0,0.3)$   & 2,708 & 5.82  & 6.45  & $+0.63$ & 20.72 \\
			$[0.3,0.5)$ & 1,130 & 40.36 & 41.78 & $+1.43$ & 56.11 \\
			$[0.5,0.7)$ & 1,900 & 61.10 & 64.92 & $+3.81$ & 65.68 \\
			$[0.7,0.9)$ & 5,234 & 82.21 & 85.15 & $+2.94$ & 75.56 \\
			$[0.9,1.0]$ & 5,187 & 94.60 & 95.11 & $+0.51$ & 69.98 \\
			\bottomrule
		\end{tabular}
	\end{table}

	\textbf{6) Ablation of hyperparameter sensitivity.} Table~\ref{tab:hyperparameter_ablation} presents a sensitivity analysis of key hyperparameters including $K$, $\tau$, $\lambda_g$ and $\gamma_{\mathrm{dlg}}$ in DLG, and $\gamma_{\mathrm{lcr}}$ in LCR. Among the tested settings, $K=5$ achieves the best results across all four metrics. Using three candidates may omit plausible referents, whereas increasing the number to seven slightly degrades performance, possibly by introducing additional competing regions. The area threshold $\tau=0.02$ yields the highest mIoU and precision scores, although $\tau=0.05$ gives a slightly higher gIoU. A smaller threshold limits the number of samples benefiting from DLG, while a larger threshold includes more large targets with limited referent ambiguity. Such samples may weaken the emphasis on candidate competition during ranker training. Increasing $\lambda_g$ from 0.5 to 1.0 degrades all metrics, suggesting that geometry is most useful as a softly weighted auxiliary cue for resolving semantically similar candidates. Similarly, stronger prior guidance at $\gamma_{\mathrm{dlg}}=1.0$ reduces performance and may amplify erroneous spatial injection while weakening boundary details. For LCR, $\gamma_{\mathrm{lcr}}=1.0$ achieves the best results, with Pr@0.9 decreasing from 40.10\% to 38.71\% and 39.19\% at 0.5 and 1.5 respectively. This pattern is consistent with insufficient correction at low strength and excessive boundary adjustment at high strength.

	\begin{table}[htbp]
		\captionsetup{justification=centering}
		\caption{Hyperparameter sensitivity analysis on RISBench.}
		\label{tab:hyperparameter_ablation}
		\centering
		\footnotesize
		
		\renewcommand{\arraystretch}{1.18}
		\setlength{\tabcolsep}{2.0pt}
		\begin{tabular}{
			>{\centering\arraybackslash}p{2.6cm}
			>{\centering\arraybackslash}p{1.0cm}
			>{\centering\arraybackslash}p{1.05cm}
			>{\centering\arraybackslash}p{1.05cm}
			>{\centering\arraybackslash}p{1.05cm}
			>{\centering\arraybackslash}p{1.05cm}}
			\toprule
			\multirow{2}{*}{\textbf{Parameter}} &
			\multirow{2}{*}{\textbf{Value}} &
			\multicolumn{4}{c}{\textbf{Performance (\%)}} \\
			\cmidrule(lr){3-6}
			& & \textbf{Pr@0.5} & \textbf{Pr@0.9} & \textbf{mIoU} & \textbf{gIoU} \\
			\midrule
			\multirow{3}{*}{\makecell{Candidate\\number ($K$)}}
			& 3 & 77.16 & 40.05 & 69.93 & 75.52 \\
			& \textbf{5} & \textbf{77.28} & \textbf{40.10} & \textbf{70.10} & \textbf{75.53} \\
			& 7 & 77.13 & 40.08 & 70.02 & 75.50 \\
			\cmidrule(lr){1-6}
			\multirow{3}{*}{\makecell{Foreground area\\threshold ($\tau$)}}
			& 0.01 & 76.88 & 39.94 & 69.81 & 75.48 \\
			& \textbf{0.02} & \textbf{77.28} & \textbf{40.10} & \textbf{70.10} & 75.53 \\
			& 0.05 & 77.07 & 39.82 & 69.96 & \textbf{75.55} \\
			\cmidrule(lr){1-6}
			\multirow{2}{*}{\makecell{Geometric correction\\weight ($\lambda_g$)}}
			& \textbf{0.5} & \textbf{77.28} & \textbf{40.10} & \textbf{70.10} & \textbf{75.53} \\
			& 1.0 & 76.91 & 40.03 & 69.89 & 75.44 \\
			\cmidrule(lr){1-6}
			\multirow{2}{*}{\makecell{Prior guidance\\strength ($\gamma_{\mathrm{dlg}}$)}}
			& \textbf{0.5} & \textbf{77.28} & \textbf{40.10} & \textbf{70.10} & \textbf{75.53} \\
			& 1.0 & 77.17 & 39.93 & 69.88 & 75.48 \\
			\midrule
			\multirow{3}{*}{\makecell{Residual combination\\coefficient ($\gamma_{\mathrm{lcr}}$)}}
			& 0.5 & 77.14 & 38.71 & 69.69 & 75.26 \\
			& \textbf{1.0} & \textbf{77.28} & \textbf{40.10} & \textbf{70.10} & \textbf{75.53} \\
			& 1.5 & 76.83 & 39.19 & 69.60 & 75.34 \\
			\bottomrule
		\end{tabular}
	\end{table}

	\subsection{Limitations} \label{sec:limitations}
\begin{figure}[htb]
		\vspace{-4pt}
		\captionsetup{belowskip=-2pt}
		\centering
		\includegraphics[width=\linewidth]{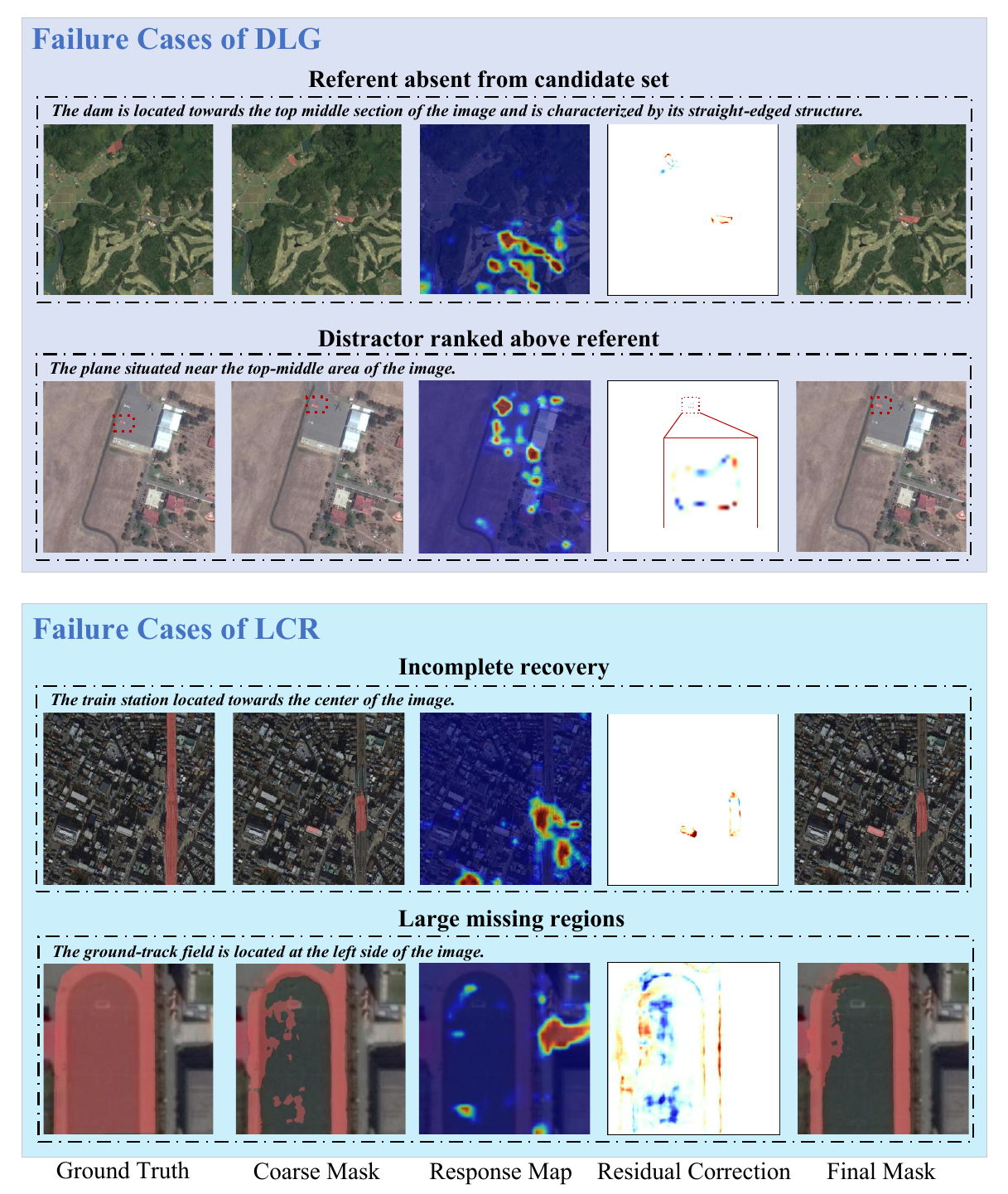}
		\caption{Representative failure cases of DiCoR. }
		\label{fig:limitation_visualization}
	\end{figure}

	Despite its overall effectiveness, DiCoR exhibits distinct limitations in referent localization and contour refinement, as illustrated in Fig.~\ref{fig:limitation_visualization}. For DLG, successful disambiguation requires both candidate coverage and accurate ranking. When the true referent produces an insufficiently salient response, it may be excluded from the top-$K$ candidates, leaving the ranker without a valid target hypothesis. Even when a valid candidate is retained, visually similar distractors may receive higher scores and lead to incorrect spatial guidance. Heavy occlusion or ambiguous linguistic cues may further impair candidate generation and ranking by weakening the visual evidence or its alignment with the expression. As DLG operates on existing multimodal features, it has limited ability to remedy deficiencies in early feature extraction and cross-modal fusion. For LCR, refinement is conditioned on the coarse prediction and trained with an emphasis on contour neighborhoods. This design favors local boundary adjustments but is less effective at recovering long missing contours or reconstructing large absent regions. As illustrated by the train station and ground-track field examples, insufficient coarse foreground support can limit LCR's recovery of extensive missing structures.
	
	\section{Conclusion} \label{sec:conclusion}
	In this paper, we presented DiCoR, a decoupled referent disambiguation and contour recalibration framework for efficient referring remote sensing image segmentation. DiCoR addresses the accuracy-efficiency dilemma between joint fusion segmentation and decoupled prompt segmentation by preserving an efficient inference pipeline while introducing dedicated mechanisms for improving grounding reliability and contour recovery. Specifically, the proposed disambiguation-aware localization guidance strategy organizes response-based candidate regions and ranks them with adaptive linguistic features, enabling the model to identify the referred target under SAM-assisted distractor supervision. In addition, the lightweight contour recalibration module learns residual boundary corrections from diverse coarse predictions under localized contour supervision, thereby improving mask delineation without relying on a foundation model during inference. Extensive experiments on RefSegRS, RRSIS-D, and RISBench demonstrate that DiCoR consistently improves segmentation accuracy and achieves a favorable accuracy-efficiency trade-off compared with existing RRSIS methods. Future work will further explore more robust early-stage localization and global boundary modeling to handle more challenging referential ambiguity and large-scale target structures.
	
	\bibliography{IEEEabrv, bib/paper}
	\bibliographystyle{IEEEtran}
	
	
	%
	%
	%
	%

	\vfill

\end{document}